\documentclass{article}

\usepackage{arxiv}
\usepackage[utf8]{inputenc} 
\usepackage[T1]{fontenc}    
\usepackage{hyperref}       
\usepackage{url}            
\usepackage{booktabs}       
\usepackage{amsfonts}       
\usepackage{nicefrac}       
\usepackage{microtype}      
\usepackage{lipsum}		

\usepackage{graphicx}   
\usepackage{subfigure}
\usepackage{caption}  
\usepackage{amsmath}
\usepackage{lineno,hyperref}
\usepackage{multirow}

\title{Deep Learning in current Neuroimaging: a multivariate approach with power and type I error control but arguable generalization ability}

\author{
  C. Jiménez-Mesa\thanks{carmenj@ugr.es} \\
  Department of Signal Theory and Communications\\
  University of Granada\\
  Granada, Spain \\
   \And
  J. Ramírez\thanks{javierrp@ugr.es} \\
  Department of Signal Theory and Communications\\
  University of Granada\\
  Granada, Spain \\
   \And
 J. Suckling\thanks{js369@cam.ac.uk} \\
  Department of Psychiatry\\
  University of Cambridge\\
  Cambridge, UK\\
  \texttt{jg825@cam.ac.uk} \\
       \And
   J. Vöglein\thanks{Jonathan.Voeglein@med.uni-muenchen.de} \\
   Department of Neurology \\
   Ludwig-Maximilians-Universität München \\ 
   Munich, Germany \\
   \\
   German Center for Neurodegenerative Diseases (DZNE)\\
   Munich, Germany \\
     \And
   J. Levin\thanks{Johannes.Levin@med.uni-muenchen.de} \\
   Department of Neurology \\
   Ludwig-Maximilians-Universität München \\ 
   Munich, Germany \\
   \\
   German Center for Neurodegenerative Diseases (DZNE)\\
   Munich, Germany \\
     \And
  J.M. Gorriz\thanks{gorriz@ugr.es, jg528@cam.ac.uk} \\
  Department of Signal Theory and Communications\\
  University of Granada\\
  Granada, Spain \\
  \\
   Department of Psychiatry\\
  University of Cambridge\\
  Cambridge, UK\\
       \And
       International Initiatives\\
for the Alzheimer's Disease Neuroimaging Initiative (ADNI)\thanks{Data used in preparation of this article were obtained from the Alzheimer’s Disease Neuroimaging Initiative (ADNI) database (adni.loni.usc.edu). As such, the investigators within the ADNI contributed to the design and implementation of ADNI and/or provided data but did not participate in analysis or writing of this report. A complete listing of ADNI investigators can be found at: \url{http://adni.loni.usc.edu/wp-content/uploads/how_to_apply/ADNI_Acknowledgement_List.pdf}}\\
for the Dominantly Inherited Alzheimer Network (DIAN) \\
}

\begin{document}
\maketitle

\begin{abstract}
Discriminative analysis in neuroimaging by means of deep/machine learning techniques is usually tested with validation techniques such as K-fold, hold-out, and resubstitution, whereas the associated statistical significance remains largely under-developed due to their computational complexity. In this work, a non-parametric framework is proposed that estimates the statistical significance of classifications using deep learning architectures that are increasingly used in neuroimaging. In particular, a combination of autoencoders (AE) and support vector machines (SVM) is applied to: (i) a one-condition, within-group designs often of normal controls (NC) and; (ii) a two-condition, between-group designs which contrast, for example,  Alzheimer's disease (AD) patients with NC (the extension to multi-class analyses is also included). A random-effects inference based on a label permutation test is proposed in both studies using cross-validation (CV) and resubstitution with upper bound correction (RUB) as validation methods. This allows both false positives and classifier overfitting to be detected as well as estimating the statistical power of the test. Several experiments were carried out using the Alzheimer’s Disease Neuroimaging Initiative (ADNI) dataset, the Dominantly Inherited Alzheimer Network (DIAN) dataset, and a MCI prediction dataset. We found in the permutation test that CV and RUB methods offer a false positive rate close to the significance level and an acceptable statistical power (although lower using cross-validation). A large separation between training and test accuracies using CV was observed, especially in one-condition designs. This implies a low generalization ability as the model fitted in training is not informative with respect to the test set. We propose as solution by applying RUB, whereby similar results are obtained to those of the CV test set, but considering the whole set and with a lower computational cost per iteration. 
\end{abstract}


\section{Introduction}\label{sec:introduction}
Machine learning (ML) assisted diagnosis of Alzheimer Disease (AD) is a key application due to its potential for a significant impact on the social well-being and health of our ageing societies. A successful system is likely to be strongly grounded in medical imaging techniques, such as Magnetic Resonance Imaging (MRI) or Positron Emission Tomography (PET); non-invasive procedures that provide rich diagnostic and prognostic information. Nevertheless, we currently remain without a complete description of the underlying neuropathological processes, especially in the early stages of the disease when interventions have their greatest impact. A particularly challenging task is to distinguish between AD and its related neurological disorders, including the prodromal stage of Mild Cognitive Impairment (MCI). \cite{teipel_relative_2015,Martinez-Murcia2016,khedher_independent_2016,illan_spatial_2014,segovia_comparative_2012}. 

The implementation of ML assistance  in neuroimaging is frequently known as computer aided diagnosis (CAD) . Current CAD systems are successful, with high accuracy rates in the classification of several conditions \cite{davatzikos_machine_2019}. Recently, there has been a trend away from ML towards  deep learning (DL)  \cite{Jo2019}. These techniques are characterised by a high degree of adaptability to input data, which allows them to be applied to  highly complex problems \cite{martinez2018}. The high dimensionality of neuroimages, or indeed most medical images, can be significantly reduced by applying Convolutional Neural Networks (CNN)\cite{Ortiz2016,payan2015} that are capable of extracting simpler patterns (initial layers) before expansion to more complex patterns (final layers). The CNN already has greater prevalence than the more traditional Multi-layer Perception (MLP) \cite{Ruck1990}, and can be applied independently \cite{Wen2020} or in conjunction with other types of classifier, such as  Support Vector Machines (SVM) \cite{Feng2020}. In addition, CNNs can also be used as layers within other more complex networks such as U-net \cite{ronneberger2015u} or Autoencoder (AE) structures, widely used in neuroimaging for dimensionality reduction \cite{Martinez-Murcia2020,Basu2019,Suk2013}. It should be acknowledged that all these studies are only possible thanks to international initiatives that offer Open Access data to the neuroimaging community, such as the Alzheimer’s Disease Neuroimaging Initiative (ADNI) \cite{Weiner2013} and the Dominantly Inherited Alzheimer Network (DIAN) \cite{Morris2012}, as well as several international competitions \cite{Neu2016,Castiglioni2018} that offer databases with the aim of increasing and improving the knowledge related to a specific task.

Among the disadvantages of neural networks are the selection of the most suitable layer configurations and the complexity associated with its training phase, with concomitantly high computational cost. Furthermore, the classification performance of these networks is highly dependent on the sample size and the number of epochs used in training; if a reduced database is used or the number of epochs is very high, overfitting is likely to occur. Under optimal conditions, i.e. with a large data set, in which the difference between the training error and test (generalization) error is small, one can assume that a good system has been obtained. Nonetheless, even with corrupted data, e.g. by deliberate permutation of data labels, due to the high capacity of the neural network, the trained model may still classify elements correctly according to the new false labels  \cite{kawaguchi2017}.  Taking all these considerations into account, the capacity of these neural networks to generalize is currently under discussion, and methods to measure and detect the configurations that generalize from those that do not would be advantageous \cite{kawaguchi2017,neyshabur2017}. For example, techniques such as dropout \cite{srivastava2014}, batch normalization \cite{ioffe2015} or data augmentation \cite{Shorten2019} are increasingly applied to reduce the generalization error \cite{Thanapol2020}. This problem is particularly common with medical data as the number of samples is usually significantly lower than the dimensionality of the accompanying features. Moreover, the estimation of the classifier error through traditional approaches, such as cross-validation or leave-one-out cross-validation, which are used in conjunction with a variance-based bounds, enable calculation of confidence interval, which are not always  effective estimators of the true error for small sample sets  \cite{Golland2005, Ojala2009}. It is therefore reasonable to consider whether the results generated by classifiers in these circumstances can be accepted. 

Prior work has analysed the reliability of classification results \cite{Pereira2009, Golland2003,Rosenblatt2019}, as well as the generation of statistical maps \cite{Etzel2009,Stelzer2013,Gorriz2021} in neuroimaging. The idea of these studies is to test if the estimated accuracy is significantly better than that obtained by chance. To this end, the most widely used approaches are permutation methods \cite{Good2013}. The permutation test procedures enable calculation of a p-value that reflects the fraction of randomly labelled data sets that result in an equal or better classifier performance than using the original data. The most common and traditional permutation tests assume for evaluation of null hypotheses that there exists independence between labels and features, and therefore there is no difference between the permuted classes. Nevertheless, several studies have found that this approach leads to biased p-values, suggesting permutations of data columns instead of labels \cite{Ojala2009}, or the use of location tests instead of accuracy tests \cite{Rosenblatt2019}. According to J.D. Rosenblatt et al. \cite{Rosenblatt2019} the causes for low power of accuracy tests are related to the discrete nature of the accuracy test statistic (especially if the test statistic is highly concentrated), the inefficient use of the data,  regularization, and the type of signal accuracy tests they are designed to detect.

The proposed approach in this paper lies within a framework for statistical significance analysis known as statistical learning theory (SLT). The main idea is to analyse the generalization capacity of a given classifier by comparing the results obtained using randomly labelled data and the ground truth. A similar study \cite{Eklund2016}  focused on the estimation of true familywise error rates by analysing the null distribution of functional magnetic resonance imaging (fMRI) datasets acquired at rest in order to verify the validity of parametric fMRI analysis. Here, we focus instead on machine learning performance with deep learning architectures. In this study, two scenarios are analysed. First, a one-condition task is analysed with healthy control (HC) data and a putative task design that generates results that should control the family-wise error (FWE). Second, a two-conditions design comparing AD and HC samples is considered, as well as a four-condition study. As labels on these samples are permuted, no group detectable differences should be apparent, which is reflected in the classification accuracy. For this purpose, several experiments were conducted with  the ADNI and  DIAN Open Access databases, applying label permutation tests.

The structure chosen for the presentation of the work is as follows. First, the databases are detailed, as well as the techniques used for classification and validation, and the proposed method itself. Then, the results are shown divided into different experiments depending on the model configuration and database, which are complemented by two studies on generalisation error and the influence of the placement of feature extraction in the proposed method.

\section{Datasets}

\subsection{The ADNI dataset}
Data used in the preparation of this article were obtained from the Alzheimer’s Disease Neuroimaging Initiative (ADNI) database (adni.loni.usc.edu). The ADNI was launched in 2003 as a public-private partnership, led by Principal Investigator Michael W. Weiner, MD. The primary goal of ADNI has been to test whether serial magnetic resonance imaging (MRI), positron emission tomography (PET), other biological markers, and clinical and neuropsychological assessment can be combined to measure the progression of mild cognitive impairment (MCI) and early Alzheimer’s disease (AD). The dataset used is composed of structural magnetic resonance images (sMRI) acquired at $1.5$T from $229$ HC and $188$ AD participants. Table \ref{tab:demog} shows demographic details of the dataset.  All images were processed by realignment, coregistration and spatially normalization to the standard SPM template, using the SPM12 software \cite{Friston1994} applying default parameters (non-rigid, preserve concentrations). In addition, SPM12 was used to segment the images obtaining Grey Matter (GM) and White Matter (WM) maps \cite{Ashburner2005}. In total, $417$ GM maps of dimensions $121\times 145\times 121$ were used in this work, normalised to the intensity range $[0, 1]$.

\begin{table}[htbp]

\caption{Demographics Details of the Datasets}\label{tab:demog}
\centering
\begin{tabular}{c|cccccc}
\textbf{Datasets} & \textbf{Status} & \textbf{Number} & \textbf{Gender (M/F)} & \textbf{Age} & \textbf{MMSE} & \textbf{CDR} \\ \hline
\multirow{2}{*}{ADNI} & HC & 229 & 119/110 & 75.97 [5.00] & 29.00 [1.00] & - \\
 & AD & 188 & 99/89 & 75.36 [7.50] & 23.28 [2.00] & - \\ \hline
\multirow{4}{*}{MCI prediction} & HC & 100 & 48/52 & 73.47 [5.76] & 29.09 [1.11] & - \\
 & MCI & 100 & 51/49 & 72.27 [7.71] & 28.05 [1.73] & - \\
 & cMCI & 100 & 60/40 & 72.47 [6.89] & 27.34 [1.86] & - \\
 & AD & 100 & 52/48 & 74.10 [7.70] & 23.13 [2.10] & - \\ \hline
\multirow{2}{*}{DIAN} & NC & 123 & 49/74 & 38.13 [11.88] & - & 0.00 [0.00] \\
 & MC & 123 & 78/45 & 37.55 [9.93] & - & 0.26 [0.50]
\end{tabular}%

\medskip
Group means and their standard deviation are shown. Symbol ``-" indicates that no information is available.

\end{table}

\subsection{MCI prediction dataset}

This dataset  was made available for the ``International challenge for automated prediction of MCI from MRI data" (\url{https://inclass.kaggle.com/c/mci-prediction}) hosted on the Kaggle platform. Patients were labelled in four classes: HC participants, AD patients, MCI participants whose diagnosis did not change in the follow-up (MCI) and converter MCI (cMCI) participants that progressed from MCI to AD in the follow-up period.  Each participant has $429$ demographic, clinical as well as cortical and subcortical MRI features estimated by Freesurfer (v5.3) \cite{fischl_measuring_2000,fischl_freesurfer_2012}. For this study, the training and test datasets were merged. In total, $400$ participants compose the database. Their demographic information is shown in Table \ref{tab:demog}. Further information about the challenge is available at \ref{appe:kaggle}.

\subsection{The DIAN dataset}
The Dominantly Inherited Alzheimer Network (DIAN) dataset (dian.wustl.edu) results from an initiative to monitor the evolution of people at risk of inheriting an autosomal dominant AD mutation. Each participant had standardized longitudinal assessments which includes clinical, cognitive, neuroimaging, CSF and plasma tests. More than 450 participants are included in DIAN. Dominantly Inherited Alzheimer’s Disease (DIAD) is caused by mutations in the Amyloid Precursor Protein (APP) \cite{Levy1990}, Presenilin-1 (PSEN1) \cite{Ryan2016} (most frequently found), or Presenilin-2 (PSEN2) \cite{Levy-Lahad1995} genes. Further information on the study is given in \cite{Morris2012}.

The images used were from baseline assessments of the DIAN Observational Study Data Freeze $14$. A total of $1219$ datasets were available. Applying exclusion factors indicated in \ref{appe:dian}, the total number of datasets selected reduced to $246$, which are labelled as non-carriers (NC) and mutation carriers (MC), with 123 participants in each group. Their demographic information is shown in Table \ref{tab:demog}. Global Clinical Dementia Rating (CDR) scale data are provided \cite{Morris1997,Berg1988}, contrasting with the Mini-Mental State Examination (MMSE) scores which were acquired for the ADNI and MCI prediction datasets. The features used consisted of imaging and non-imaging biomarkers: $188$ MRI features and $520$ FDG (Fluorodeoxyglucose F18) PET features. The remaining 14 features were non-imaging biomarkers obtained from CSF and blood plasma, where protocols INNO, xMAP, PL\textunderscore xMA and Lumipulse were used for measurements \cite{Fagan2014}. Specifically, fibrillar amyloid-$\beta$ (A$\beta$) depositions and CSF $\tau$ protein were measured as markers: A$\beta _{40}$, A$\beta _{42}$, A$\beta _{40}$:A$\beta _{42}$ ratio, $\tau$, $p$-$\tau$. An APOE  genetic test was also undertaken.

\section{Methodology}
\begin{figure*}[htbp]
\centering
\includegraphics[scale=0.6]{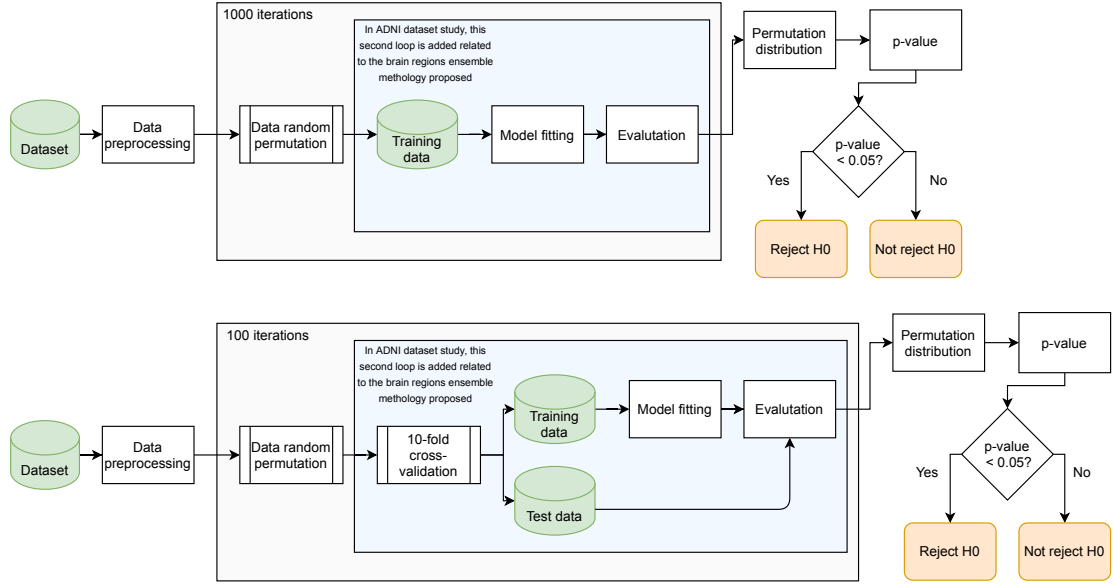}
\caption{Flowchart of the proposed method using resubstitution with upper bound correction (top) and K-fold cross-validation (bottom) for statistical power assessment and type I error control analysis.}
\label{fig:metodos}
\end{figure*}

We propose a statistical significance assessment framework for ML and DL systems: the application of permutation tests to assess type I errors and  statistical power (related to type II errors) of a classifier using upper-bounded resubstitution as the validation method. Specifically, our classifier is an AE-SVM configuration. For comparative purposes, the experiments were also conducted using stratified $10$-fold cross-validation. An overview of the pipeline is illustrated in Figure \ref{fig:metodos}. 

\subsection{Statistical analysis based on SLT}
Given  $\mathbf{S} = \lbrace \left(  x_i, y_i  \right)  \rbrace_{i=1}^n  $ as a dataset where $\mathbf{x} \in \mathbb{R}^N$ are the observed features and $\mathbf{y} \in \mathbb{R}^M$ the class labels, an analysis of differences between conditions included in a dataset can be established as a hypothesis test where the null hypothesis, $H_0$, implies that there is independence between data and labels, ${p(\mathbf{x},\mathbf{y}) = p(\mathbf{x})p(\mathbf{y})}$, or in terms of functional neuroimaging there is no ``effect" in $\mathbf{x}$ given $\mathbf{y}$. If $H_0$ is rejected then there is a dependency between data and labels, that is, the distribution $\lbrace \mathbf{x}\vert \mathbf{y} = 0 \rbrace$ is different from $\lbrace \mathbf{x}\vert \mathbf{y} = 1 \rbrace$. To perform this hypothesis test, the out-sample prediction error, or actual error, associated with classification is the chosen test statistic.

Given the dataset $\mathbf{S}$ and a learning algorithm $L$, it is possible to calculate the estimate error of the algorithm, $\hat{E}_L$.When using resubstitution with upper bound correction as the evaluation method (RUB), the estimated error is defined as:
\begin{equation}
 \hat{E}_L^{RUB} = \dfrac{1}{n} \sum_{i=1}^n I\lbrace L_{\mathbf{S}}\left( x_i  \right) \neq y_i  \rbrace + \mu
\end{equation}
where $L_S$ is the function obtained by the learning algorithm $L$ given the dataset $\mathbf{S}$, $I\left( \cdot \right) $  is the indicator function and $\mu$ represents the upper bound \cite{vapnik1982}. The estimation error is associated with the actual risk while the first term represents the empirical risk. More detailed information on the derivation of the upper bound is described in \ref{appe:bound}. An expression for the estimation of the upper bound is:
\begin{equation} \label{eq:gorriz}
\mu_{emp} \leq  \sqrt{ \dfrac{1}{2n} \ln {\dfrac{2 \sum_{k=0}^{d-1}{\left( \begin{array}{cc} n-1 \\ k \end{array}\right) } } {\eta} } }
\end{equation}
where $\eta$ is the significance level, $n$ is the size of the training set, and $d$ is the feature's dimension. It should be noted that this expression is only applicable to linear classifiers \cite{Gorriz2019}, such as SVM. 

Regarding K-fold cross-validation, each of the estimates obtained from each of the $K$ partitions, or folds, was considered independently, instead of considering the mean across partitions. Then, the estimator used is:
\begin{equation}
 \hat{E}_L^{KCV} =  \dfrac{1}{card\left( S^k\right) } \sum_{i \in S^k} I\lbrace L_{S^{(k)}}\left( x_i  \right) \neq y_i  \rbrace
\end{equation}
where the k-th partition $S^k$ is  the test set and the remainder of the data $S^{(k)}$, the training set. Thus, the test statistics are detonated as $\mathcal{T}^{RUB}$ and $\mathcal{T}^{KCV}$ for RUB and CV approaches, respectively.

As in the case of RUB, errors associated to cross-validation could be connected by an upper bound. According to the theory related to generalization errors, in the worst-case $E_{act} = \Delta + E_{emp} $, where $\Delta$ could be another representation of the upper bound. The empirical error is the error related to the training set, whilst the actual error is related to the test set. Ideally, such an upper bound would be zero, as this means that both errors are equal, and therefore, the classifier is able to generalize. In mathematical terms, this can be easily observed in the following expression:
\begin{equation}\label{eq:bound_CV}
\dfrac{\Delta }{E_{emp}} = \dfrac{E_{act}}{E_{emp}} -1
\end{equation}

Therefore, a positive $\Delta /E_{emp}$ implies a poor generalization ($E_{act} > E_{emp}$) since $\dfrac{E_{act}}{E_{emp}} \propto  \dfrac{\Delta }{E_{emp}}$, whereas if the ratio becomes negative, it would be even better than the ideal situation, as that means that the actual error is lower than the empirical error. Nevertheless, in order to be able to analyse this pattern, the bound should be constant, because if it is not, there is no effective learning during training as the two errors are not related to each other.

\subsubsection{Permuted test statistics}\label{sec:permutation}
Permutation tests enable comparison of the test statistic obtained for each dataset and its permutation distribution (null distribution). The detailed explanation of the steps involved in a permutation test for a range of classification scenarios is given in \ref{appe:perm_test}. A label-switching permutation test procedure was chosen despite its conservative performance \cite{Ojala2009}. This procedure was preferred to parametric inference because these experiments involved high dimensional databases of small sample size  (\textit{curse of dimensionality}) where central limit approximations tend to be poor \cite{Rosenblatt2019,Gorriz2021}.

The estimated test statistic obtained by permutation is denoted as $\mathcal{T_{\pi}}$. The resulting null distribution contained $1000$ of these values requiring $1000$ iterations in the case of upper-bounded implementation, while for $10$-fold validation $100$ iterations were computed (100$\times$10). Considering each fold independently allows calculation of the total variance of the method instead of the average variance generated by the $10$ folds, although its computational cost is considerably higher than applying RUB. One iteration of $10$-fold is approximately equal to $10$ iterations of RUB.  The observed test statistics, $\mathcal{T}^{RUB}$ and $\mathcal{T}^{KCV}$were derived from $20$ iterations for both approaches, with the aim of obtaining the most accurate estimation of the actual errors. At each iteration, the dataset labels were shuffled.

\subsubsection{Statistical power: stated hypothesis}
The first of the two experiments conducted on the databases assessed statistical power. The null hypothesis associated with this study, in which two-condition and four-conditions datasets were tested, is:
\begin{equation} \label{eq:hipotesis_power}
\begin{split}
& H_0 : \mathcal{T_{\pi}} = \mathcal{T}  \\
& H_1 : \mathcal{T_{\pi}} > \mathcal{T}
\end{split}
\end{equation}
Therefore, the null hypothesis assumes that the classification error associated with the permuted dataset is equal to the actual error of the original dataset, since there is independence between $\mathbf{x}$ and $\mathbf{y}$. The alternative hypothesis is that there is an ``effect" in $\mathbf{x}$ given $\mathbf{y}$, and therefore the error obtained by permutation is larger than the actual error.

To analyse the probability of observation of the null hypothesis, the mathematical expression associated with the computation of the p-value is:
\begin{equation}\label{eq:pvalue}
p_{value}  = \dfrac{card\lbrace \mathcal{T_{\pi}} \leq \mathcal{T} \rbrace +1} {M+1}
\end{equation}
where $card (.)$ denotes the cardinality. The addition of $1$ to the numerator and denominator is equivalent to including the fixed statistic value in the test, which allow to obtain a valid test \cite{Reiss2015}. 

\subsubsection{Type I error control}

The second experiment evaluated the type I error control of the inference method by testing one-condition distributions, in particular the within-group HC condition. The HC dataset was randomly split in two  groups of the same size. 

The null hypothesis of independence between samples and labels was true by construction. Thus, the proportion of false positives detected should be close to the significance level. In order to compute the proportion of analyses (or iterations) that give rise to any significant result, an Omnibus test was applied. First, a similar expression to equation (\ref{eq:pvalue}) was used for the computation of the p-values of dimension $M$:
\begin{equation}
p_{value,m} = \dfrac{card\lbrace \mathcal{T}_{\pi} \leq  \mathcal{T}_{\pi,m} \rbrace} {M}
\end{equation}
where $\mathcal{T}_{\pi,m}$ represents the m-th error from the $M$ actual errors obtained by the permutation test, which was compared to the permutation distribution of errors (including itself). Once the $M$ p-values were obtained, they were compared with the significance level $\alpha$, and the number of false positives calculated as the number of these values that were less than or equal to the significance level. The estimated FWE rate (or FP rate) was the number of false positives divided by the number of permutations:
\begin{equation}
FWE \: \: rate = \dfrac{card\lbrace p_{values} \leq  \alpha \rbrace} {M}
\end{equation}

From a parametric perspective, this analysis could be performed with a contrast of means, so that from the errors obtained a confidence interval was constructed in which $50\%$ were included, since the dataset was randomly split into two sets of the same condition.

In both experiments, the level of significance $\alpha$ was set as $0.05$. When computing p-values including the standard deviation, it was considered as the empirical p-value obtained but did not include all possible permutations of the databases, due to computational restrictions. The standard deviation related to the Monte Carlo approximation of the p-value is $\sqrt{\dfrac{p\left( 1-p\right) }{n}}$ \cite{Efron1993}, where $p$ is the p-value obtained and $n$ is the number of samples.

From these experiments it was also possible to check the classifier's capacity to correctly predict random sample labels, giving insight into the susceptibility to overfitting of the  neural networks.

\subsection{System configuration}
\subsubsection{DL and ML architectures}
\begin{figure}[htbp]
\centering
\includegraphics[scale=0.6]{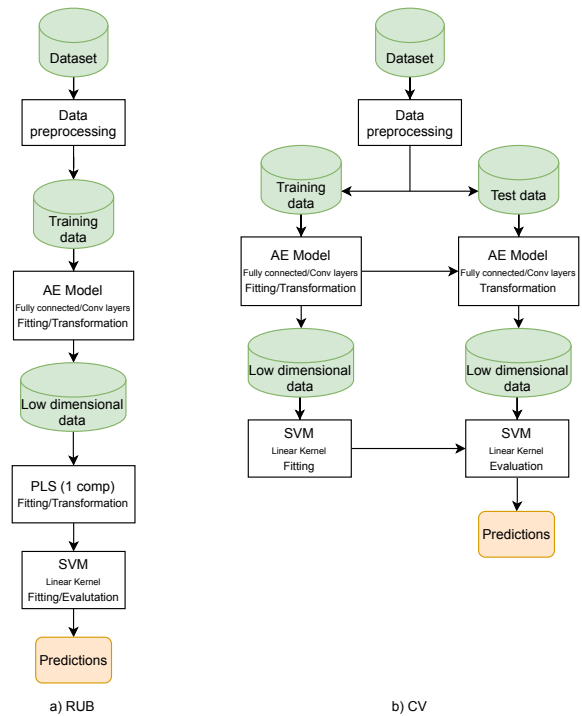}
\caption{General flowchart of feature extraction and classification methods using a) RUB validation (left) and b) $10$-fold cross-validation (right).}
\label{fig:flow_clasificacion}
\end{figure}

The system adopted for the classification phase was based on AE and SVM architectures, which are widely used in medical imaging \cite{Feng2020, Martinez-Murcia2020, Basu2019, Lopez2009, Jimenez-Mesa2020, Gorriz2017, Ghazal2018}. The AE was used for feature extraction while the SVM (linear kernel) was used as the classifier. The reasons for selecting these architectures include: i) good performance in an initial phase of searching for architectures of interest, ii) the possibility of using a DL architecture to reduce dimensionality in a low-sample size scenario, and iii) linearity in the final classification level for applying RUB. In addition to linearity, it is desirable that the number of input features to the classifier is as low as possible (ideally $1$) since there exists a high dependency of the bound  on the features dimensions. Therefore, when using RUB as a validation method  the dimensionality of the features extracted from the Z-layer was reduced by applying partial least squares (PLS) \cite{rosipal2005}. This technique allows extraction of features (or components) by combining information about the variances of the initial features and their associated labels and the correlation amongst them. Thus, it was possible to work with a lower dimensionality at the cost of assuming a decrease in the variance explained. One PLS component was used in the experiments. For a flowchart of the classification phase, see Figure \ref{fig:flow_clasificacion}.

For the design of the AE, the layers of the configuration were chosen depending on the input features.  For one-dimensional features, fully-connected layers were used for reducing the feature space dimension, whilst for 3-dimensional features (for example, GM images) convolutional layers were mostly selected instead. Sigmoid activation functions were selected for one-dimensional features since input features were  scaled, whilst rectified linear unit (ReLU) activation functions were used in hidden layers for 3-dimensional feature since images were normalised to the range $[0, 1]$. The theoretical background to this architecture can be found in \ref{appe:dl_teoria}.

The SVM directly selects the sample label when used with the DIAN and MCI  datasets. A more sophisticated design was needed in the experiments associated with the ADNI database, where the SVM was configured as an ensemble of several brain regions instead of the whole brain. Further explanation about the regions is explained in Section \ref{sec:exp_adni}. To obtain the label for a sample $y$ , posterior probabilities returned by SVM are necessary, so that for each sample the total probability of belonging to a class, $C$, depends on the $L$ regions:
\begin{equation}
p_{total}^c\left( y \right) = \sum_l \dfrac{p_l^c\left( y \right)}{n_{regions}}
\end{equation}
where $p_{total}^c\left( y \right)$ is the probability of a sample $y$  being of class $c$, and the term $p_l^c\left( y \right)$ is the probability of a sample $y$ being of class $c$ in a specific region $l$. Then, the maximum value among the probabilities for each class is chosen as a label:
\begin{equation} \label{eq:probabilidad}
Label(y) = \underset{y}{\arg \max } \left( p_{total}^c\left( y \right) \right) 
\end{equation}

\subsubsection{Autoencoder settings}
\begin{figure}[!h]
\centering
\subfigure[The ADNI]{\label{fig:ae_adni}\includegraphics[width=100mm]{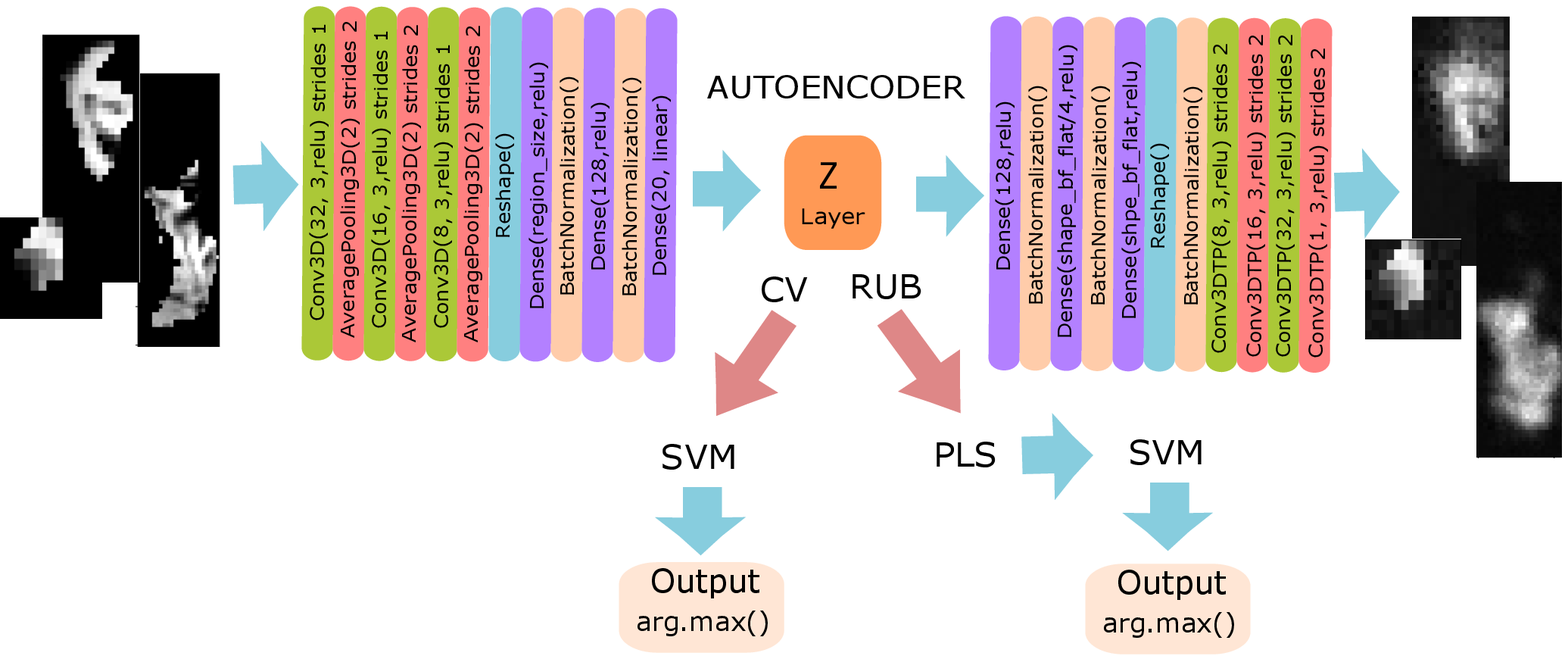}} \\
\subfigure[MCI prediction]{\label{fig:ae_kaggle}\includegraphics[width=60mm]{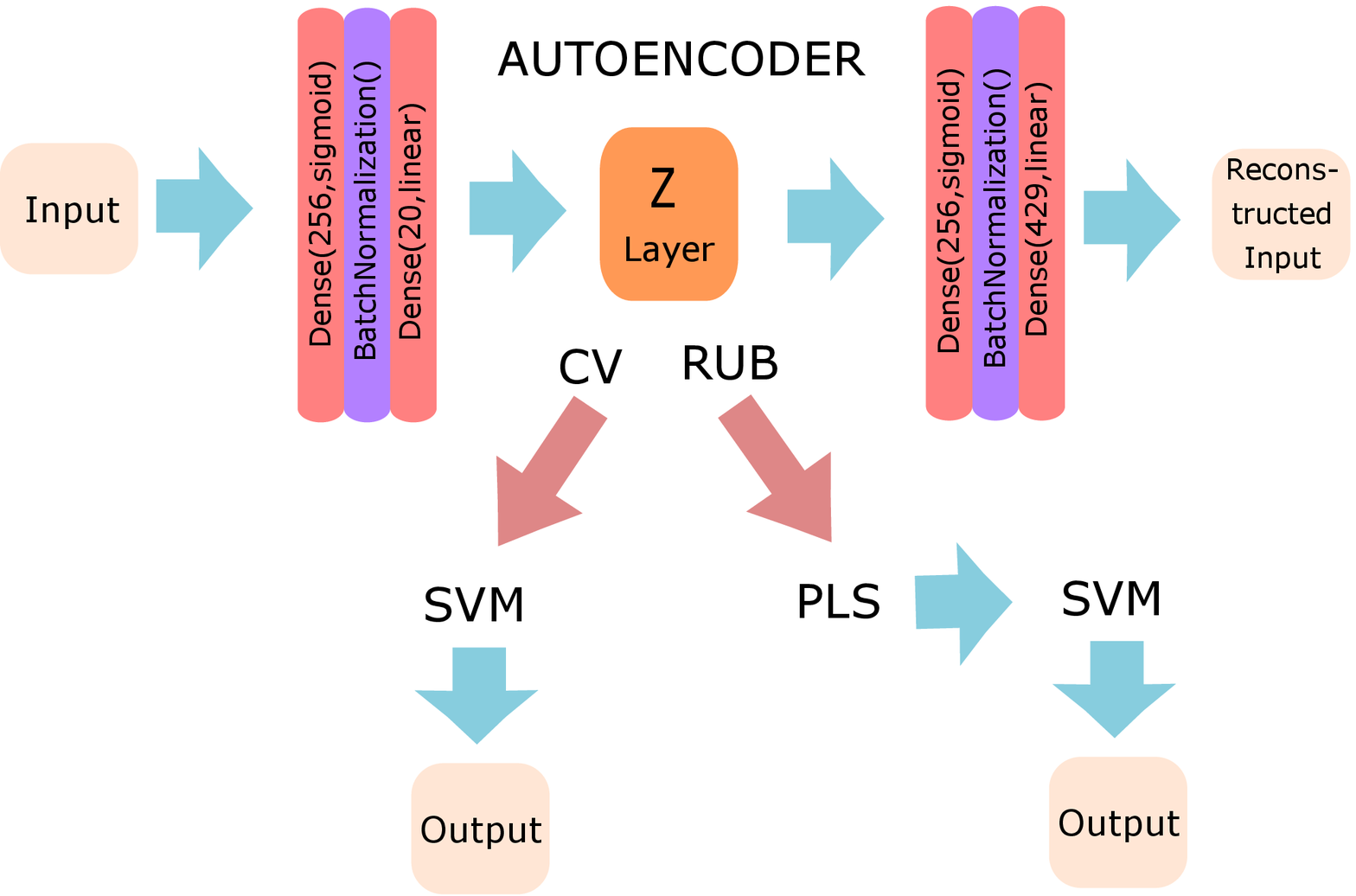}}
\subfigure[The DIAN]{\label{fig:ae_dian}\includegraphics[width=60mm]{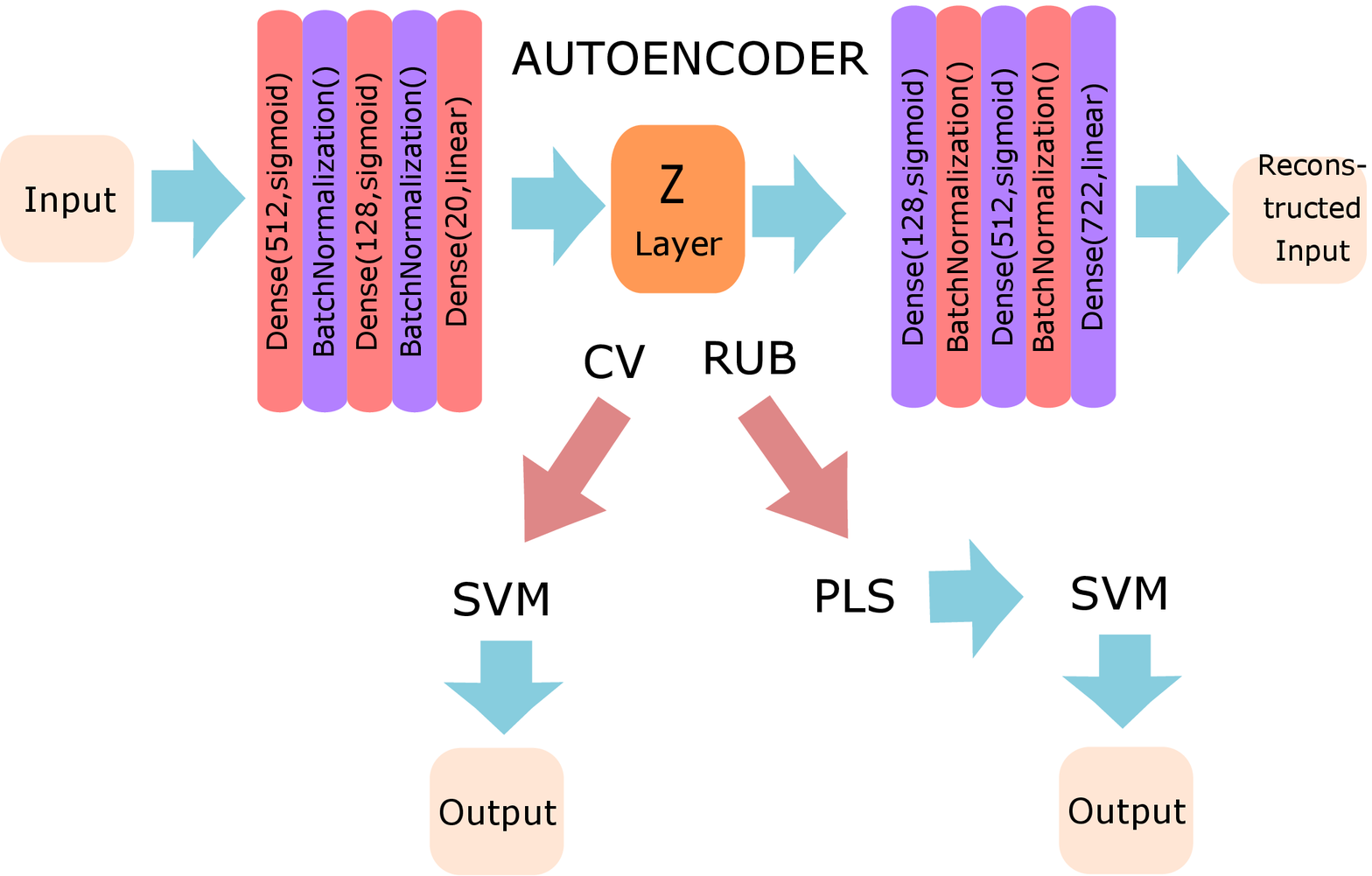}}

\caption{Model (AE - SVM) configuration for (a) the ADNI dataset. The images shown as input to the model are examples of axial slices of the 3D used images, (b) MCI prediction dataset and (c) the DIAN dataset.}
\label{fig:ae_models}
\end{figure}

Prior to the experiments described in this paper, a stratified $5$-fold cross-validated grid-search was done to identify configurations (layer typology, number of layers, and hyperparameters) that work effectively with the original (non-permuted) databases. The  features  from each dataset were reduced to $20$  in the Z-layer. These features were introduced as input to the SVM linear classifier in the experiments that used $10$-fold CV or were reduced to $1$ feature by using PLS and then introduced to the classifier where RUB was applied. Figure \ref{fig:ae_models} illustrates the layer configurations related to each dataset. Batch normalization \cite{ioffe2015} was applied in all AEs to reduce overfitting.

The optimizer applied in the experiments with the three datasets was Adam \cite{kingma2019} with a default learning rate  of $0.001$ for MCI prediction and DIAN experiments, whilst the learning rate was $0.0001$ in the ADNI experiment. In all configurations, mean square error (MSE) was the loss function. The largest number of epochs used in training was $250$ for the ADNI dataset, which was chosen as a balance between loss separability of the classes and computational cost. For MCI prediction, the number of epochs was $50$, which was selected after observing the curve of training and validation loss, as a higher number of epochs resulted in overfitting. The number of epochs in the AE when processing the DIAN dataset was $80$, a value chosen after analysing the fit of the model to the data, again avoiding overfitting. In the training step, $30\%$ of the training data samples were used as a validation set during AE tuning. These configurations and hyperparameters were maintained in all experiments, whether the data was the original or permuted.

\section{Results}
For reasons of clarity, the results obtained for each dataset will be indicated separately as experiments. Each experiment is composed of the two already-mentioned studies: statistical power assessment and type I error control analysis. The main objective of these experiments is not to obtain high classification accuracy rates but analyse the trade-off between statistical power and type I error. Two other studies complementary to the above are presented: i) to analyse the performance of cross-validation in terms of upper bounds, and ii) to apply feature extraction outside the permutation test to analyse variability in the results.

The scripts generated for this study are based on Python language (v.3.7) and Keras library over Tensorflow on a Nvidia Tesla V100-SXM2.

\subsection{Experiment 1: ADNI dataset} \label{sec:exp_adni}
\begin{figure}[!h]
\centering
\subfigure[Selected regions]{\label{fig:regiones_atlas}\includegraphics[width=80mm]{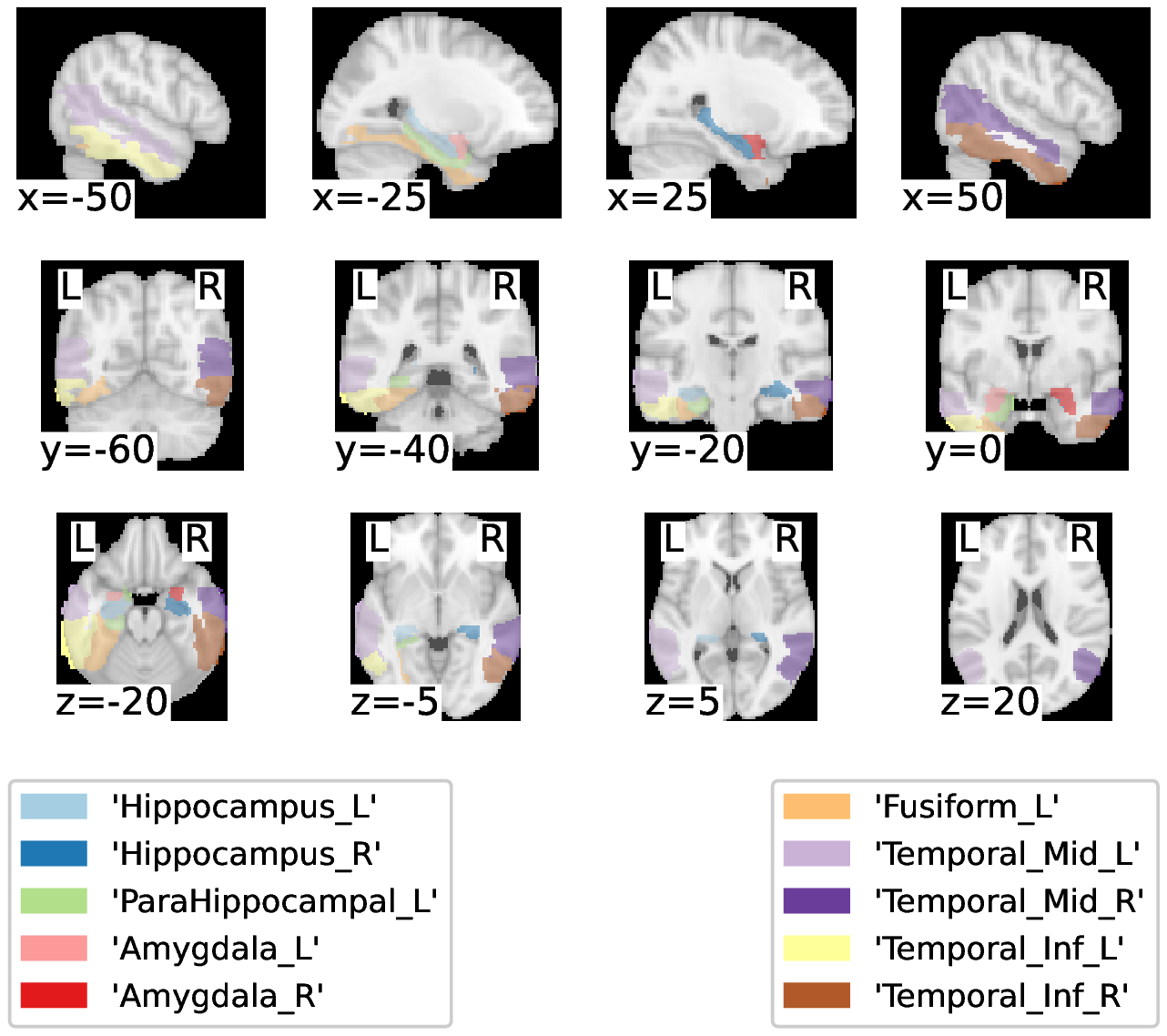}}
\subfigure[$10$-fold CV accuracies]{\label{fig:fold_region}\includegraphics[width=80mm]{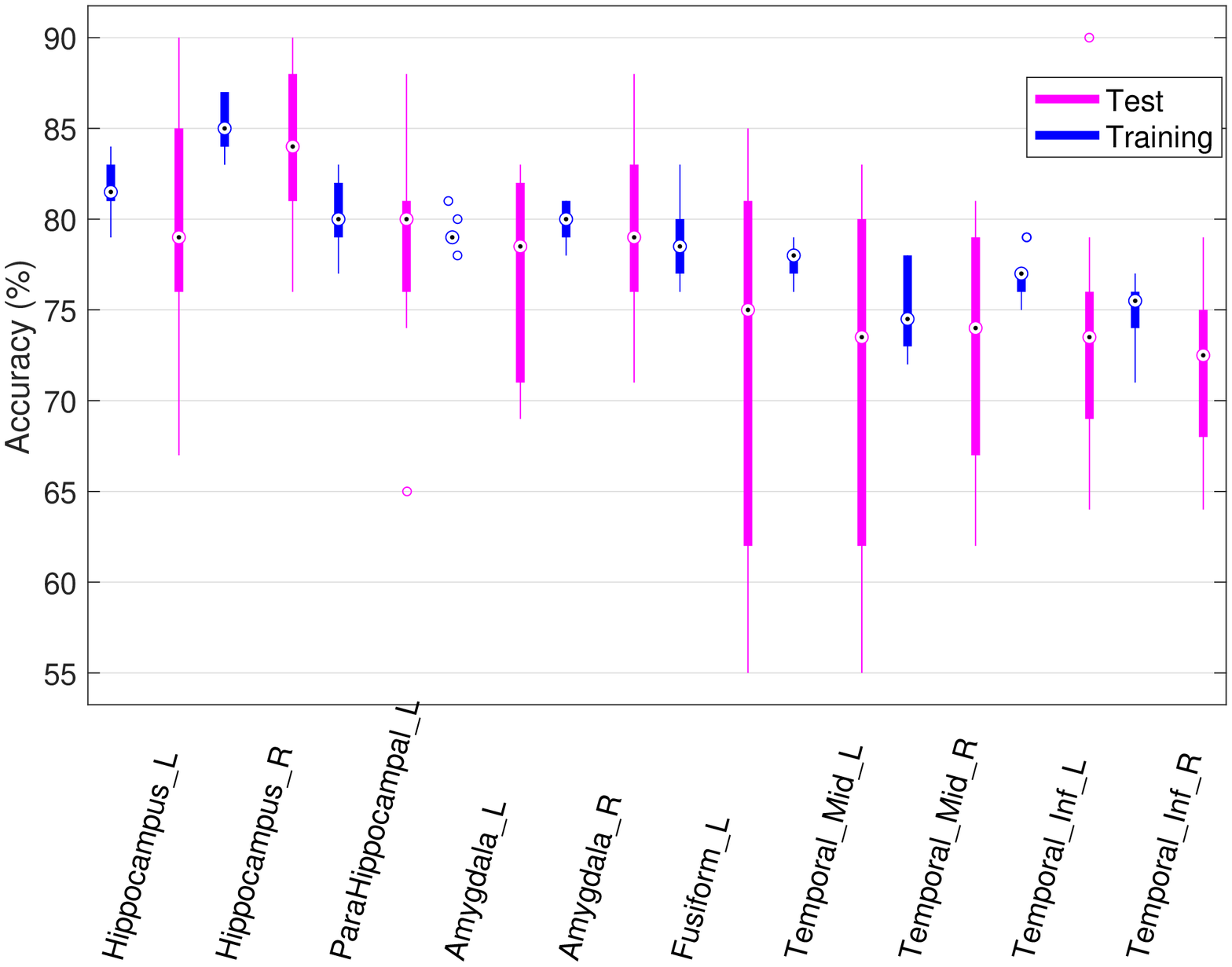}}

\caption{(a) Selected regions from AAL atlas. (b) Training and test accuracies of each region applying $10$-fold cross-validation to the ADNI original dataset.}
\end{figure}

This experiment is a two-group (AD vs HC) classification of three-dimensional features; that is, GM maps derived from sMRI images. The whole brain image was not used, but instead $10$ specific regions of the Automated Anatomical Labeling (AAL) atlas \cite{Tzourio-Mazoyer2002}: hippocampus (left/right), parahippocampal (left), amygdala (left/right), fusiform (left), middle temporal gyrus (left/right) and inferior temporal gyrus (left/right). These regions were selected according to their importance to AD diagnosis \cite{Gorriz2021} and are shown in Figure \ref{fig:regiones_atlas}. The use of a limited number of regions reduces the computational cost  making the ensemble methodology tractable, and the results more reliable overall. The architecture was trained using each region as an input, and once the a posteriori probabilities related to each region were obtained, equation (\ref{eq:probabilidad}) was applied to get the final labels. 

Following the framework shown in Figure \ref{fig:flow_clasificacion}, $417$ subjects of the ADNI  dataset were used to train and test the model proposed, and is illustrated in Figure \ref{fig:ae_adni} for $20$ iterations. The results of this procedure are shown in Table \ref{tab:adni-real} in the ``Original dataset" rows. An example of accuracy rates related to each region is shown in Figure \ref{fig:fold_region} for one $10$-fold cross-validation iteration. Thereafter, the scheme illustrated in Figure \ref{fig:metodos} was applied to perform the permutation tests.

\begin{table*}[!h]

\caption{ADNI. Results Related to the Statistical Power Experiment}
\label{tab:adni-real}
\centering
\begin{tabular}{cccccc}
 &  & \textbf{Accuracy} & \textbf{Sensitivity} & \textbf{Specifity} & \textbf{p-value} \\ \hline
\multicolumn{6}{l}{\textbf{10-fold CV}} \\ \hline
\multirow{2}{*}{Original dataset} & Training & 0.8548 [0.0094] & 0.9166 [0.0094] & 0.7794 [0.0163] &  - \\ \cline{2-6} 
 & Test & 0.8319 [0.0541] & 0.8989 [0.0627] & 0.7504 [0.0932] & - \\ \hline
\multirow{2}{*}{Permuted dataset} & Training & 0.4791 [0.0864] & 0.4359 [0.0993] & 0.5219 [0.1026]  & - \\ \cline{2-6} 
 & Test & 0.4987 [0.0753] & 0.4550 [0.1187] & 0.5425 [0.1180]  & 0.0010 [0.0010]  \\ \hline
\multicolumn{6}{l}{\textbf{Resubstitution}} \\ \hline
\multicolumn{2}{c}{Original dataset} & 0.8321 [0.0059] & 0.9033 [0.0075] & 0.7455 [0.0117] & - \\ \hline
\multicolumn{2}{c}{Permuted  dataset} &  0.5380 [0.0339] & 0.4854 [0.1608] &  0.5904 [0.1539] & 0.0010 [0.0010]   \\ \hline
\multicolumn{6}{l}{\textbf{Upper-bounded resubstitution ($\mu = 0.0665$)}} \\ \hline
\multicolumn{2}{c}{Original dataset} & 0.7656 [0.0059] & $\ast$ & $\ast$  & - \\ \hline
\multicolumn{2}{c}{Permuted  dataset} &  0.4715 [0.0339] & $\ast$ & $\ast$ &  0.0010 [0.0010]  
\end{tabular}%

\medskip
Mean scores and their standard deviations (in squared brackets) related to the ADNI original and permuted datasets. Symbols ``-" and $\ast$ indicate that values were not computable or values were obtained from the resubstitution method, respectively.  Validation methods applied to the permuted dataset were $10$-fold cross-validation ($100$ iterations, high computational cost, top), resubstitution ($1000$ iterations, high computational cost, middle) and upper-bounded resubstitution (by applying the upper bound, low computational cost, bottom). Original dataset scores were obtained from $20$ iterations (medium computational cost). The significance level of the test was $0.05$.
\end{table*}

\subsubsection{Randomization on HC vs AD }

All sample labels in the ADNI dataset were permuted prior to  fitting and evaluation of the model. To apply the RUB approach it was necessary to calculate the upper bound associated with each experiment by means of equation (\ref{eq:gorriz}). An alternative expression for computing the bound is equation (\ref{eq:vapnik}). The number of participants included was $417$, the learning rate, $\eta$, was $0.05$, and the number of features introduced into the SVM classifier was $1$ after applying PLS to the $20$ features of the AE Z-Layer. Therefore, the value of the upper bound in this case is $0.0665$ using equation (\ref{eq:gorriz})  and $0.2103$ using equation (\ref{eq:vapnik}). These values demonstrate that the upper bound obtained from equation (\ref{eq:vapnik}) is more conservative, as mentioned in \ref{appe:bound}. For this reason, only the results obtained using equation (\ref{eq:gorriz}) are presented here.

\begin{figure}[!h]
\centering
\includegraphics[scale=0.4]{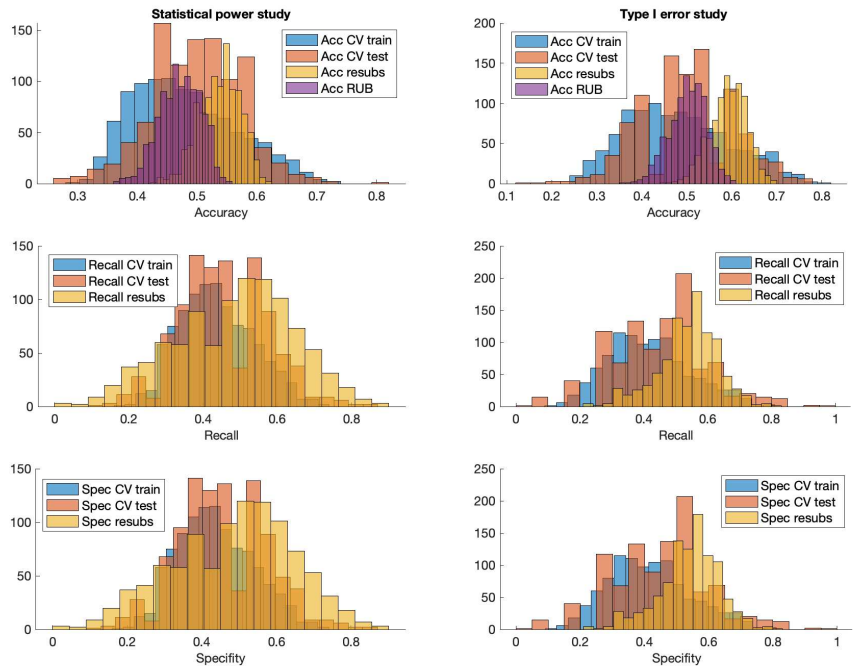}
\caption{Distribution of scores of the permuted ADNI dataset in the statistical power study (left) and type I error study (right).}
\label{fig:hist_adni}
\end{figure}

The results of the permutation test are shown in Table \ref{tab:adni-real}. The actual errors (i.e. test statistics) from the original dataset were $\mathcal{T}^{KCV}=0.1681$ and $\mathcal{T}^{RUB}=0.2344$ using $10$-fold CV and upper-bounded resubstitution as validation methods, respectively. Table \ref{tab:adni-real} shows that these test statistics are considerably lower than the mean errors associated with the permutation distribution ($0.5013$ and $0.5285$). Moreover, Figure \ref{fig:hist_adni} (left) illustrates the distribution of the several metrics reported in Table  \ref{tab:adni-real} over the iterations of the permutation test. It should be remembered that although $100$ iterations were performed with the $10$-fold method, there were actually $1000$ values when all the folds are considered together. With this information, the p-values obtained are equal to $0.0010$ for both cross-validation and upper-bounded resubstitution method. Indeed, no error belonging to the null distribution was lower than the statistic test in any case. As both p-values were smaller than the significance level imposed in the test ($0.05$), the null hypothesis was rejected. Therefore, there was no independence between conditions (HC, AD) and samples.

\subsubsection{Randomization on HC}

In this experiment, a subset of the HC sample was used to evaluate the type I error control of the model. The sample labels were permuted and inputted into the model, as in the previous experiment. The number of samples was $229$ and thus the value of the bound was $0.0897$ using equation~(\ref{eq:gorriz}). The distribution of the scores of the permutation test is shown in Figure \ref{fig:hist_adni} (right). A summary of the results is indicated in Table \ref{tab:adni-hc}, with mean actual errors of $0.4988$ and $0.4966$ using $10$-fold CV and RUB, respectively. The FWE rates obtained were $0.0410$ using CV and $0.0440$ using RUB. Hence, the number of false positives obtained was consistent with the level of significance, $0.05$.

\begin{table*}[!h]

\caption{ADNI. Results Related to the Type I Error Experiment}
\label{tab:adni-hc}
\centering
\begin{tabular}{cccccc}
 &  & \textbf{Accuracy} & \textbf{Sensitivity} & \textbf{Specifity} & \textbf{FWE rate} \\ \hline
\multicolumn{6}{l}{\textbf{10-fold CV}} \\ \hline
\multirow{2}{*}{Permuted dataset} & Training & 0.4790 [0.1152] & 0.4151 [0.1273] & 0.5423 [0.1259] &  - \\ \cline{2-6} 
 & Test & 0.5012 [0.1007] & 0.4455 [0.1484] & 0.5568 [0.1497] & 0.0410 [0.0063] \\ \hline
\multicolumn{6}{l}{\textbf{Resubstitution}} \\ \hline
\multicolumn{2}{c}{Permuted  dataset} &  0.5931 [0.0406] & 0.5390 [0.0921] & 0.6466 [0.0865] & 0.0440 [0.0065]  \\ \hline
\multicolumn{6}{l}{\textbf{Upper-bounded resubstitution ($\mu = 0.0897$)}} \\ \hline
\multicolumn{2}{c}{Permuted  dataset} & 0.5034 [0.0406]  & $\ast$ & $\ast$ & 0.0440 [0.0065]
\end{tabular}%

\medskip
Mean scores and their standard deviations (in squared brackets)  related to the ADNI permuted HC subset. Symbols ``-" and $\ast$ indicate that values were not computable or values are related to resubstitution method, respectively.  Validation methods applied to the permuted dataset were $10$-fold cross-validation ($100$ iterations, high computational cost, top), resubstitution ($1000$ iterations, high computational cost, middle) and upper-bounded resubstitution (by applying the upper bound, low computational cost, bottom). The significance level of the test was $0.05$.
\end{table*}

\subsection{Experiment 2: MCI prediction dataset}
This experiment was conducted from three perspectives: two-condition (AD vs HC) and four-condition (AD vs cMCI vs MCI vs HC) permutation distributions were used for statistical power assessment, in addition to the estimation of type error I control with HC subjects. The configuration of the model did not vary between using four or two conditions in the analysis since one vs one methodology was used \cite{Jimenez-Mesa2020}. Comparison with the challenge results indicate that the chosen settings had an accuracy rate on the original no-dummies test subset of $0.44 [0.04]$ after shuffling the data two times.

\subsubsection{Randomization on HC vs AD }

\begin{figure*}[!h]
\centering
\includegraphics[scale=0.40]{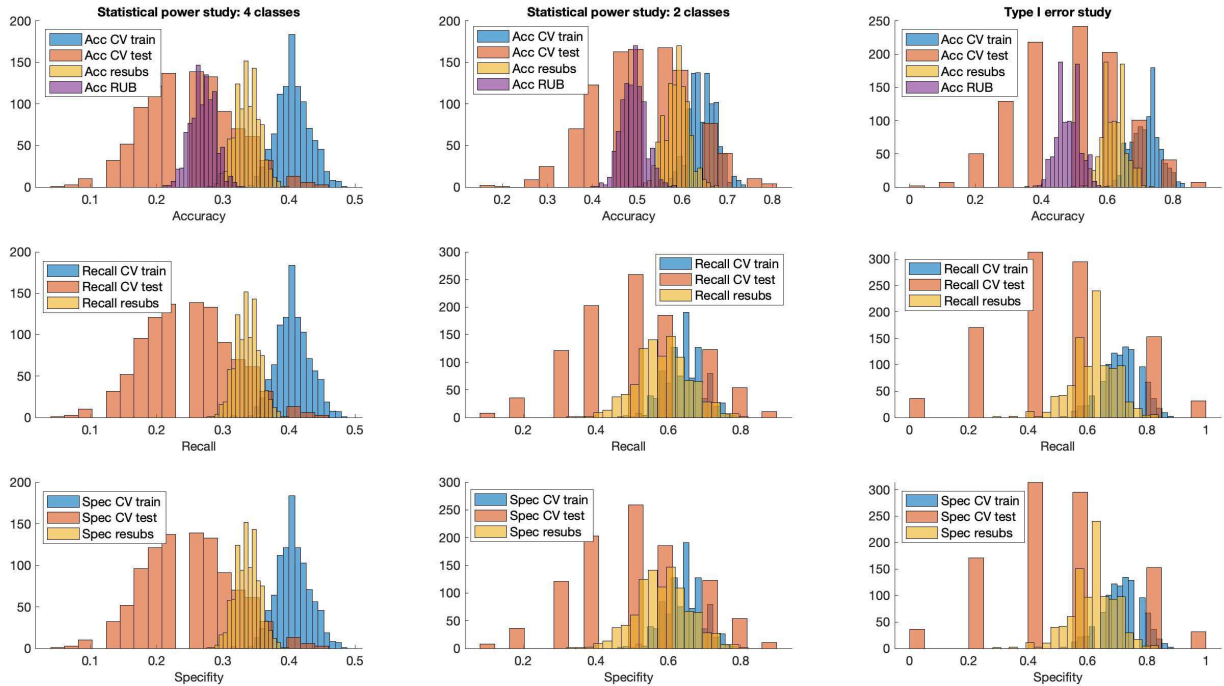}
\caption{Distribution of scores of the permuted MCI prediction dataset  in the statistical power study for four classes (left), two classes (middle) and type I error study (right).}
\label{fig:hist_kaggle}
\end{figure*}

Table \ref{tab:kaggle-real} summarizes the results for the permutation test using $10$-fold CV and RUB with the upper bound correction in the four-condition analysis. Computing the upper bound, $n=400$, $d=1$ and $\eta = 0.05$, which applying equation (\ref{eq:gorriz}) gives $\mu = 0.0679$. In summary, the test statistics were $\mathcal{T}^{KCV} = 0.5604$ and $\mathcal{T}^{RUB} = 0.6142$. Figure \ref{fig:hist_kaggle} (left) illustrates the distribution of the scores of the permuted dataset. In this case, the obtained p-values were  $0.0040$ and $0.0010$ for CV and RUB, respectively. Therefore, the null hypothesis of independence between labels and samples was rejected with both methods. 

\begin{table*}[!h]

\caption{MCI Prediction (Multiclass). Results Related to the Statistical Power Experiment}
\label{tab:kaggle-real}
\centering
\begin{tabular}{ccccc}
 &  & \textbf{Accuracy} & \textbf{Sensitivity} & \textbf{p-value} \\ \hline
\multicolumn{5}{l}{\textbf{10-fold CV}} \\ \hline
\multirow{2}{*}{Original dataset} & Training & 0.5823 [0.0071] & 0.5823 [0.0071] & - \\ \cline{2-5} 
 & Test & 0.4396 [0.0181] & 0.4396 [0.0181] & - \\ \hline
\multirow{2}{*}{Permuted dataset} & Training & 0.4066 [0.0250] & 0.4066 [0.0250] & - \\ \cline{2-5} 
 & Test & 0.2487 [0.0687] & 0.2487 [0.0687] & 0.0040 [0.0020]\\ \hline
\multicolumn{5}{l}{\textbf{Resubstitution}} \\ \hline
\multicolumn{2}{c}{Original dataset} & 0.4537 [0.0162] & 0.4537 [0.0162]  & - \\ \hline
\multicolumn{2}{c}{Permuted  dataset} & 0.3375 [0.0184] & 0.3375 [0.0184] & 0.0010 [0.0010] \\ \hline
\multicolumn{5}{l}{\textbf{Upper-bounded resubstitution ($\mu = 0.0679$)}} \\ \hline
\multicolumn{2}{c}{Original dataset} & 0.3858 [0.0162] & $\ast$ & - \\ \hline
\multicolumn{2}{c}{Permuted  dataset} & 0.2696 [0.0184] & $\ast$ & 0.0010 [0.0010]
\end{tabular}%

\medskip
Mean scores and their standard deviations (in squared brackets) related to MCI prediction original and permuted dataset (4 classes). Symbols ``-" and $\ast$ indicate that values were not computable or values are related to resubstitution method, respectively.  Validation methods applied for the permuted dataset were $10$-fold cross-validation ($100$ iterations, medium computational cost, top), resubstitution ($1000$ iterations, medium computational cost, middle) and upper-bounded resubstitution (by applying the upper bound, low computational cost, bottom). Original dataset scores were obtained from $20$ iterations (low computational cost). The significance level of the test was $0.05$.
\end{table*}

Table \ref{tab:kaggle-real-binario} summarizes the results for  the statistical power assessment in the two-condition distribution. In this case, the actual errors associated to the original dataset were $\mathcal{T}^{KCV} = 0.1542$ and $\mathcal{T}^{RUB} = 0.2798$ and the mean actual error of the null distributions were $0.4933$  using $10$-fold CV and $0.5087$ with RUB. The values used  for computation of the upper bound were the same as those for the four-condition analysis, except the number of samples, which was $200$ in this case. Thus, the value of the upper bound was $0.0960$. Figure \ref{fig:hist_kaggle} (middle) shows the distribution of the accuracies and  other metrics obtained in this analysis. In this study, p-values were $0.0010$ in both cases, CV and RUB. Thus, there was an ``effect" in the samples given the labels.

\begin{table*}[!h]

\caption{MCI Prediction (Binary). Results Related to the Statistical Power Experiment}
\label{tab:kaggle-real-binario}
\centering
\begin{tabular}{cccccc}
 &  & \textbf{Accuracy} & \textbf{Sensitivity} & \textbf{Specifity} & \textbf{p-value} \\ \hline
\multicolumn{6}{l}{\textbf{10-fold CV}} \\ \hline
\multirow{2}{*}{Original dataset} & Training & 0.9141 [0.0158] & 0.9288 [0.0227] & 0.8994 [0.0224] & - \\ \cline{2-6} 
 & Test & 0.8458 [0.0775] & 0.8510 [0.1084] & 0.8405 [0.1165] & - \\ \hline
\multirow{2}{*}{Permuted dataset} & Training & 0.6455 [0.0302] & 0.6441 [0.0527] & 0.6468 [0.0534] & - \\ \cline{2-6} 
 & Test & 0.5067 [0.1061] & 0.5052 [0.1574] & 0.5083 [0.1625] & 0.0010 [0.0010] \\ \hline
\multicolumn{6}{l}{\textbf{Resubstitution}} \\ \hline
\multicolumn{2}{c}{Original dataset} & 0.8162 [0.0167] & 0.8540 [0.0216] & 0.7785 [0.0283] & - \\ \hline
\multicolumn{2}{c}{Permuted  dataset} & 0.5873 [0.0263] & 0.5868 [0.0728] & 0.5877 [0.0723] & 0.0010 [0.0010]  \\ \hline
\multicolumn{6}{l}{\textbf{Upper-bounded resubstitution ($\mu = 0.0960$)}} \\ \hline
\multicolumn{2}{c}{Original dataset} & 0.7202 [0.0167] & $\ast$ & $\ast$  & - \\ \hline
\multicolumn{2}{c}{Permuted  dataset} & 0.4913 [0.0263] & $\ast$ & $\ast$ &  0.0010 [0.0010] 
\end{tabular}%

\medskip
Mean scores and their standard deviations (in squared brackets) related to MCI prediction original and permuted AD-HC subset. Symbols ``-" and $\ast$ indicate that values were not computable or values are related to resubstitution method, respectively.  Validation methods applied for the permuted dataset were $10$-fold cross-validation ($100$ iterations, medium-low computational cost, top), resubstitution ($1000$ iterations, medium-low computational cost, middle) and upper-bounded resubstitution (by applying the upper bound, low computational cost, bottom). Original dataset scores were obtained from $20$ iterations (low computational cost). The significance level of the test was $0.05$.
\end{table*}

\subsubsection{Randomization on controls  }

Finally, a HC subset (combining HC training and test subjects) was used for type I error control estimation. Results related to this permutation test are shown in Table \ref{tab:kaggle-hc} whilst its distribution is illustrated in Figure \ref{fig:hist_kaggle} (right). In this case the number of samples was $100$, so the upper bound was equal to $0.0960$. The mean actual errors obtained in this case were $0.5108$ using $10$-fold CV and $0.5171$ using RUB. From these null distributions, the FWE rate obtained was $0.0480$ applying CV,   close to the significance level $\alpha=0.05$, and $0.0330$ applying RUB.

\begin{table*}[!h]

\caption{MCI Prediction. Results Related to the Type I Error Experiment}
\label{tab:kaggle-hc}
\centering
\begin{tabular}{cccccc}
 &  & \textbf{Accuracy} & \textbf{Sensitivity} & \textbf{Specifity} & \textbf{FWE rate} \\ \hline
\multicolumn{6}{l}{\textbf{10-fold CV}} \\ \hline
\multirow{2}{*}{Permuted dataset} & Training & 0.7175 [0.0428] & 0.7172 [0.0650] & 0.7179 [0.0667] & - \\ \cline{2-6} 
 & Test & 0.4892 [0.1557] & 0.4902 [0.2279] & 0.4882 [0.2344] & 0.0480 [0.0068] \\ \hline
\multicolumn{6}{l}{\textbf{Resubstitution}} \\ \hline
\multicolumn{2}{c}{Permuted  dataset} & 0.6187 [0.0356] & 0.6196 [0.0733] & 0.6178 [0.0765] &  0.0330 [0.0056] \\ \hline
\multicolumn{6}{l}{\textbf{Upper-bounded resubstitution ($\mu = 0.1358$)}} \\ \hline
\multicolumn{2}{c}{Permuted  dataset} & 0.4829 [0.0356] & $\ast$ & $\ast$ & 0.0330 [0.0056]
\end{tabular}%

\medskip
Mean scores and their standard deviations (in squared brackets) related to MCI prediction permuted HC subset. Symbols ``-" and $\ast$ indicate that values were not computable or values are related to resubstitution method, respectively.  Validation methods applied for the permuted dataset were $10$-fold cross-validation ($100$ iterations, medium-low computational cost, top), resubstitution ($1000$ iterations, medium-low computational cost, middle) and upper-bounded resubstitution (by applying the upper bound, low computational cost, bottom). The significance level of the test was $0.05$.

\end{table*}

\subsection{Experiment 3: DIAN dataset}
This experiment used the DIAN database ($246$ samples, $722$ features), which was conceived for prediction of the evolution of DIAD, as well as to analyse the genetic subgroups stratified by specific mutations. For these purposes, the most commonly used techniques are based on machine learning algorithms \cite{Castillo-Barnes2020, Franzmeier2020}, although studies using artificial neural networks have also been published \cite{Luckett2021}.

\subsubsection{Randomization on MC vs NC }

\begin{figure}[!h]
\centering
\includegraphics[scale=0.4]{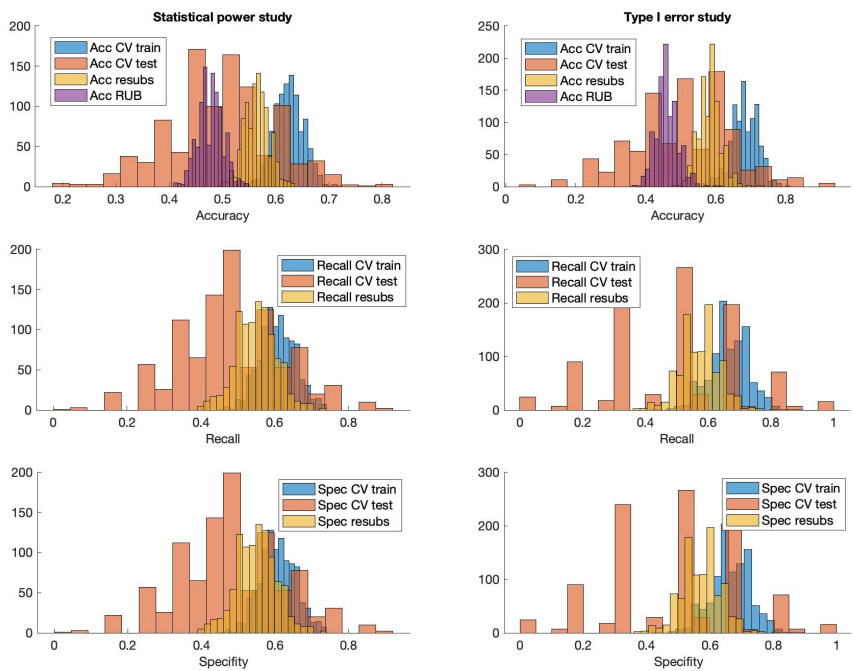}
\caption{Distribution of scores of the permuted DIAN dataset  in the statistical power study (left) and type I error study (right).}
\label{fig:hist_dian}
\end{figure}

Following the framework, assessing statistical power began by using the original database to train and test the model to obtain the test statistics. Mean results based on $20$ iterations (with data shuffled on each iteration) are shown in ``Original dataset'' rows of Table \ref{tab:dian_real}. The test statistics, i.e. the mean actual errors, were $\mathcal{T}^{KCV} = 0.3714$ and $\mathcal{T}^{RUB} = 0.4565$. To compute the null distribution, all samples were permuted and shuffled for $100$ iterations ($100\times 10$ values) when using CV and $1000$ iterations when using RUB. The upper bound used in conjunction with resusbtitution was calculated with a sample size of $246$) the number of features as input in the classifier =$1$, and the level of significance =$0.05$, giving a value of $0.0866$.  The metrics generated by the permutation test using a stratified $10$-fold cross-validation are shown in the first rows of Table \ref{tab:dian_real}, with a corresponding p-value of $0.0819$. Using the upper-bounded resubstitution (bottom of Table \ref{tab:dian_real}), the p-value was $0.0020$. Therefore, the null hypothesis was rejected (data and labels are dependent) when tested with RUB, but not CV. Figure \ref{fig:hist_dian} (left) illustrates the distribution of the several metrics reported in Table  \ref{tab:dian_real} over the iterations of the permutation test. 

\begin{table*}[!h]

\caption{DIAN. Results Related to the Statistical Power Experiment}
\label{tab:dian_real}
\centering
\begin{tabular}{cccccc}
 &  & \textbf{Accuracy} & \textbf{Sensitivity} & \textbf{Specifity} & \textbf{p-value} \\ \hline
\multicolumn{6}{l}{\textbf{10-fold CV}} \\ \hline
\multirow{2}{*}{Original dataset} & Training & 0.7325 [0.0225] & 0.7901 [0.0364] & 0.6748 [0.0359] &  - \\ \cline{2-6} 
 & Test & 0.6286 [0.0962] & 0.6802 [0.1360] & 0.5777 [0.1560] & - \\ \hline
\multirow{2}{*}{Permuted dataset} & Training & 0.6260 [0.0271] & 0.6078 [0.0483] & 0.6442 [0.0497] & - \\ \cline{2-6} 
 & Test & 0.4963 [0.0975] & 0.4785 [0.1433] & 0.5141 [0.1477] & 0.0819 [0.0087] \\ \hline
\multicolumn{6}{l}{\textbf{Resubstitution}} \\ \hline
\multicolumn{2}{c}{Original dataset} & 0.6301 [0.0355] & 0.7301 [0.0693] & 0.5301 [0.0552] & - \\ \hline
\multicolumn{2}{c}{Permuted  dataset} & 0.5641 [0.0228] & 0.5500 [0.0538] & 0.5783 [0.0529] & 0.0020  [0.0014] \\ \hline
\multicolumn{6}{l}{\textbf{Upper-bounded resubstitution ($\mu = 0.0866$)}} \\ \hline
\multicolumn{2}{c}{Original dataset} & 0.5435 [0.0355] &  $\ast$ &  $\ast$ & - \\ \hline
\multicolumn{2}{c}{Permuted  dataset} & 0.4775 [0.0228] &  $\ast$ &  $\ast$  & 0.0020 [0.0014]
\end{tabular}%

\medskip
Mean scores and their standard deviations (in squared brackets) related to the DIAN original and permuted dataset. Symbols ``-" and $\ast$ indicate that values were not computable or values are related to resubstitution method, respectively.  Validation methods applied for the permuted dataset were $10$-fold cross-validation ($100$ iterations, medium-low computational cost, top), resubstitution ($1000$ iterations, medium-low computational cost, middle) and upper-bounded resubstitution (by applying the upper bound, low computational cost, bottom). Original dataset scores were obtained from $20$ iterations (low computational cost). The significance level of the test was $0.05$.
\end{table*}

\subsubsection{Randomization on controls}

Results of the permutation test to assess type I error control are given in Table \ref{tab:dian-nc}, whilst in Figure \ref{fig:hist_dian} (right) the distribution of  several metrics is illustrated. In this case, the model was fitted and evaluated using HC. The number of participants was $123$, but the parameters remained otherwise unchanged from the preceding experiment leading to an upper bound of $0.1225$. The mean actual error related to $10$-fold CV was $0.4972$ after $100$ iterations ($100\times 10$ values) while $0.5394$ was the value associated to RUB after $1000$ iterations. The p-values obtained were $0.0300$ for CV and $0.0440$ for RUB, again p-values were close or equal to $\alpha$ ($0.05$). Therefore, the number of false positives was the number expected at that significance level.

\begin{table*}[!h]

\caption{DIAN. Results Related to the Type I Error Experiment}
\label{tab:dian-nc}
\centering
\begin{tabular}{cccccc}
 &  & \textbf{Accuracy} & \textbf{Sensitivity} & \textbf{Specifity} & \textbf{FWE rate} \\ \hline
\multicolumn{6}{l}{\textbf{10-fold CV}} \\ \hline
\multirow{2}{*}{Permuted dataset} & Training & 0.6830 [0.0378] & 0.6645 [0.0620] & 0.7013 [0.0616] & - \\ \cline{2-6} 
 & Test & 0.5028 [0.1395] & 0.4799 [0.2070] & 0.5260 [0.2101] & 0.0300 [0.0054]\\ \hline
\multicolumn{6}{l}{\textbf{Resubstitution}} \\ \hline
\multicolumn{2}{c}{Permuted  dataset} & 0.5831 [0.0314] & 0.5707 [0.0611] & 0.5953 [0.0592] & 0.0440 [0.0065] \\ \hline
\multicolumn{6}{l}{\textbf{Upper-bounded resubstitution ($\mu = 0.1225$)}} \\ \hline
\multicolumn{2}{c}{Permuted  dataset} & 0.4606 [0.0314] & $\ast$ & $\ast$ & 0.0440 [0.0065] 
\end{tabular}%

\medskip
Mean scores and their standard deviations (in squared brackets)  related to the DIAN permuted NC subset. Symbols ``-" and $\ast$ indicate that values were not computable or values are related to resubstitution method, respectively.  Validation methods applied for the permuted dataset were $10$-fold cross-validation ($100$ iterations, medium-low computational cost, top), resubstitution ($1000$ iterations, medium-low computational cost, middle) and upper-bounded resubstitution (by applying the upper bound, low computational cost, bottom). The significance level of the test was $0.05$.
\end{table*}

\subsection{Upper bounds in cross-validation: overfitting detectors}
\begin{figure}[htbp]
\centering
\includegraphics[scale=0.35]{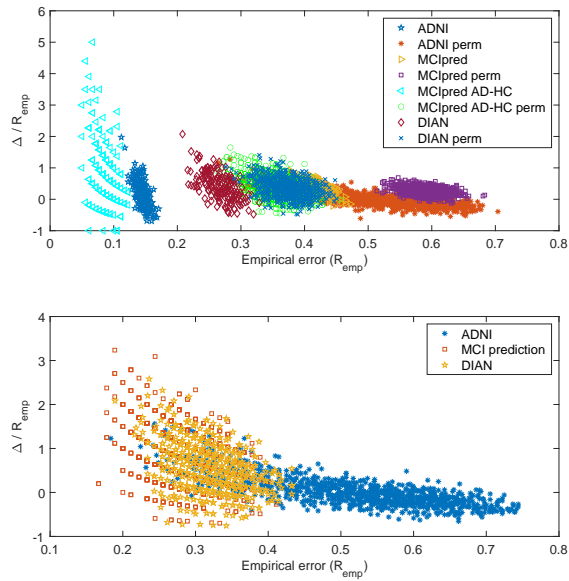}
\caption{Ratio upper bound-empirical risk, $\Delta/R_{emp}$,  versus empirical risk using $10$-fold cross-validation in the original datasets ($20$ iterations, $200$ values) and permuted datasets ($100$ iterations, $1000$ values). Values related to statistical power assessment (top) and type I error control (bottom) studies.}
\label{fig:distribuciones_kfold}
\end{figure}

A more in-depth study can be conducted with the results obtained from K-fold as validation method in order to detect overfitting. For this, equation (\ref{eq:bound_CV}) was applied. The capacity of generalisation can be analysed by studying the value and variability of the ratio $\Delta/R_{emp}$ associated with each of the iterations performed, both in original and permuted databases using $10$-fold CV. The upper graph in Figure \ref{fig:distribuciones_kfold} is a comparison of the $\Delta/R_{emp}$ ratios vs the empirical risk obtained for the original and permuted datasets in the statistical power experiment. The lower graph shows the equivalence in the experiment related to false positives control.

With regard to the upper graph, it can be seen that the the lowest empirical errors were associated to the non-permuted binary databases, which is already known from the Tables \ref{tab:adni-real}, \ref{tab:kaggle-real-binario} and \ref{tab:dian_real}. Furthermore, the ADNI ratios were closer to zero and negative values, which means that empirical and actual errors were close to each other, which coincides with that indicated in Table \ref{tab:adni-real}. Regarding the binary MCI prediction dataset, a large variation in the value of the ratio $\Delta /E_{emp}$ was observed, which in accordance with the high accuracies related to the original dataset may indicate the existence of overfitting. In the permuted databases, the cloud of points associated with the ADNI dataset is particularly noteworthy with ratios that encompass negative values. This implies that the actual errors were lower than the empirical errors, and therefore the implemented model had good generalization with permuted data. The same situation occurred in the subset of ADNI controls in the bottom graph, where it was observed that most iterations had negative values of the ratio $\Delta /E_{emp}$. With respect to the subsets of controls in the other two databases, their ratios were mostly positive, so it was not possible to declare generalizability. 

\subsection{Variability in feature extraction}\label{sec:reducida}
\begin{figure}[htbp]
\centering
\includegraphics[scale=0.60]{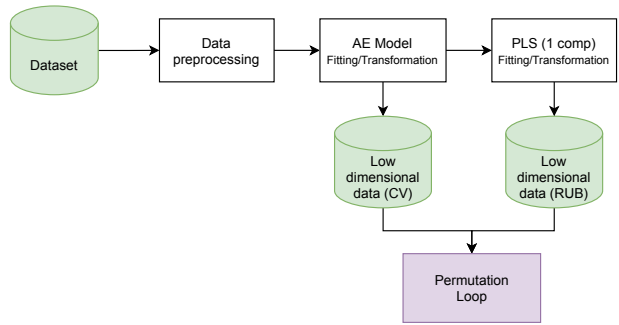}
\caption{Flowchart of feature extraction outside the permutation loop.}
\label{fig:extraccion}
\end{figure}

To study how feature extraction was affected by using a permuted database, a different approach than that illustrated in Figure \ref{fig:flow_clasificacion} was followed. In this case, feature extraction was performed only once on the original database. Thus, the permutation test was applied directly to the low dimensionality data as is illustrated in Figure \ref{fig:extraccion}. This implies that the permutation test only affected SVM classification, rather than feature extraction and classification as in framework proposed. For this study, the same criteria and parameters were applied as in the previous studies, the only difference being the use of principal component analysis (PCA) \cite{Abdi2010} instead of PLS for dimensionality reduction in the subset of control subjects, as PLS could not be applied using a single class. The aim of this experiment was to observe how the variability of low-dimensional features was affected by extracting them using the permuted databases.

Table \ref{tab:power_svm} and Table \ref{tab:fwe_svm} indicate the results obtained following this new scheme. In general, it was observed that the results associated with the permutations were effectively centred on the $50\%$ of accuracy. This is especially obvious when resubstitution was applied. In addition, the associated standard deviation  also decreased in most cases. This is consistent with the fact that by applying feature extraction to the original database and subsequently working in low dimensionality, feature extraction was not influenced by false labels and samples being shuffled, thus decreasing variability. Statistical power and FWE rate were both similar to those obtained in the prior experiments.

\begin{table*}[htbp]

\caption{Statistical Power Assessment Using the Alternative Scheme}
\label{tab:power_svm}
\centering
\resizebox{\textwidth}{!}{%
\begin{tabular}{cccccccccc}
 &  & \multicolumn{2}{c}{\textbf{ADNI}} & \multicolumn{2}{c}{\textbf{MCI Prediction (multiclass)}} & \multicolumn{2}{c}{\textbf{MCI Prediction (binary)}} & \multicolumn{2}{c}{\textbf{DIAN}} \\
 &  & \textbf{Accuracy} & \textbf{p-value} & \textbf{Accuracy} & \textbf{p-value} & \textbf{Accuracy} & \textbf{p-value} & \textbf{Accuracy} & \textbf{p-value} \\ \hline
\multicolumn{10}{l}{\textbf{10-fold CV}} \\ \hline
\multirow{2}{*}{Original dataset} & Training & 0.8611 [0.0063] & - & 0.5904 [0.0150] & - & 0.9109 [0.0129] & - & 0.7157 [0.0166] & - \\ \cline{2-10} 
 & Test & 0.8341 [0.0535] & - & 0.4581 [0.0721] & - & 0.8540 [0.0741] & - & 0.6148 [0.0948] & - \\ \hline
\multirow{2}{*}{Permuted dataset} & Training & 0.4898 [0.0822] & - & 0.4001 [0.0208] & - & 0.6456 [0.0314] & - & 0.6272 [0.0273] & - \\ \cline{2-10} 
 & Test & 0.4730 [0.0762] & 0.0010 [0.0010] & 0.2469 [0.0673] & 0.0040 [0.0020] & 0.5015 [0.1102] & 0.0010 [0.0010] & 0.4974 [0.0977] & 0.1179 [0.0102] \\ \hline
\multicolumn{10}{l}{\textbf{Resubstitution}} \\ \hline
\multicolumn{2}{c}{Original dataset} & 0.8369 [0.0000] & - & 0.4273 [0.0021] & - & 0.8100 [0.0000] & - & 0.6545 [0.0000] & - \\ \hline
\multicolumn{2}{c}{Permuted  dataset} & 0.4918 [0.0244]  & 0.0010 [0.0010] & 0.2714 [0.0153] & 0.0010 [0.0010] & 0.5247 [0.0271] & 0.0010 [0.0010] & 0.5177 [0.0217] & 0.0010 [0.0010] \\ \hline
\multicolumn{2}{l}{\textbf{RUB}} & \multicolumn{2}{c}{$\mu = 0.0665$} & \multicolumn{2}{c}{$\mu = 0.0679$} & \multicolumn{2}{c}{$\mu = 0.0960$} & \multicolumn{2}{c}{$\mu = 0.0866$} \\ \hline
\multicolumn{2}{c}{Original dataset} & 0.7704 [0.0000] & - & 0.3594 [0.0021] & - & 0.7140 [0.0000] & - & 0.5679 [0.0000] & - \\ \hline
\multicolumn{2}{c}{Permuted  dataset} & 0.4253 [0.0244] & 0.0010 [0.0010] & 0.2035 [0.0153] & 0.0010 [0.0010] & 0.4287 [0.0271] & 0.0010 [0.0010] & 0.4311 [0.0217] & 0.0010 [0.0010]
\end{tabular}%
}

\medskip
Mean values and their standard deviations (in squared brackets) of accuracy and p-values. Symbol ``-" indicates that values were not computable. Validation methods applied for the permuted dataset were $10$-fold cross-validation ($100$ iterations, medium computational cost, top), resubstitution ($1000$ iterations, low computational cost, middle) and upper-bounded resubstitution (by applying the upper bound, low computational cost, bottom). Original dataset scores were obtained from $20$ iterations (low computational cost). The significance level of the test was $0.05$.
\end{table*}

\begin{table*}[htbp]

\caption{Type I Error Assessment Using the Alternative Scheme}
\label{tab:fwe_svm}
\centering
\resizebox{\textwidth}{!}{%
\begin{tabular}{cccccccc}
 &  & \multicolumn{2}{c}{\textbf{ADNI}} & \multicolumn{2}{c}{\textbf{MCI Prediction}} & \multicolumn{2}{c}{\textbf{DIAN}} \\
 &  & \textbf{Accuracy} & \textbf{FWE rate} & \textbf{Accuracy} & \textbf{FWE rate} & \textbf{Accuracy} & \textbf{FWE rate} \\ \hline
\multicolumn{8}{l}{\textbf{10-fold CV}} \\ \hline
\multirow{2}{*}{Permuted dataset} & Training & 0.5064 [0.0924] & - & 0.7178 [0.0419] & - & 0.6893 [0.0376] & - \\ \cline{2-8} 
 & Test & 0.4267 [0.1177] & 0.0450 [0.0066] & 0.4949 [0.1517] & 0.0390 [0.0061] & 0.4939 [0.1422] & 0.0150 [0.0038] \\ \hline
\multicolumn{8}{l}{\textbf{Resubstitution}} \\ \hline
\multicolumn{2}{c}{Permuted  dataset} & 0.4882 [0.0328]  & 0.0490 [0.0068]  & 0.5324 [0.0353] & 0.0390 [0.0061] & 0.5299 [0.0343] & 0.0340 [0.0057] \\ \hline
\multicolumn{2}{l}{\textbf{RUB}} & \multicolumn{2}{c}{$\mu = 0.0897$} & \multicolumn{2}{c}{$\mu = 0.1358$} & \multicolumn{2}{c}{$\mu = 0.1225$} \\ \hline
\multicolumn{2}{c}{Permuted  dataset} & 0.3524 [0.0328] & 0.0490 [0.0068] & 0.4099 [0.0353] & 0.0390 [0.0061] & 0.4074 [0.0343] & 0.0340 [0.0057]
\end{tabular}%
}
\medskip
Mean values and their standard deviations (in squared brackets) of accuracy and p-values. Symbol ``-" indicates that values were not computable. Validation methods applied for the permuted dataset were $10$-fold cross-validation ($100$ iterations, medium computational cost, top), resubstitution ($1000$ iterations, low computational cost, middle) and upper-bounded resubstitution (by applying the upper bound, low computational cost, bottom). The significance level of the test was $0.05$.
\end{table*}

\section{Discussion}
\begin{figure}[htbp]
\centering
\includegraphics[scale=0.30]{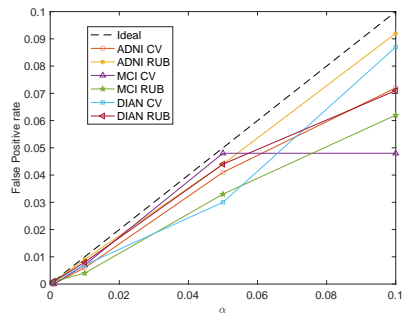}
\caption{Estimated FP rate derived from the Omnibus test of the analysed methods at given significance levels ($\alpha$).}
\label{fig:fwe_rate}
\end{figure}

The results obtained demonstrate high statistical power and good control of type I error in the experiments performed. In summary, results of the statistical analysis performed are indicated in Table \ref{tab:p-values}, where p-values and FWE rates related to Vapnik's upper bound correction are also included (see \ref{appe:bound} for its computation). Furthermore, the trend of the false positive (FP)  rate is plotted in Figure \ref{fig:fwe_rate} for the assessment of the type I error control. In it, one can observe that the FP rate is always lower than the line of identity and that a performance closer to the ideal occurs using RUB. 

\begin{table*}[!h]

\caption{Summary of the Obtained P-Values (Statistical Power Study) and FWE Rates (Type~I Error Study) }
\label{tab:p-values}
\centering
\resizebox{\textwidth}{!}{%
\begin{tabular}{ccccccc}
  & \multicolumn{3}{c}{\textbf{Permutation test: statistical power}} & \multicolumn{3}{c}{\textbf{Permutation test: type I error}} \\
  & 10-fold & RUB & R-Vapnik & 10-fold & RUB & R-Vapnik \\
\textbf{ADNI} & 0.0010 [0.0010] & 0.0010 [0.0010]  & 0.0010 [0.0010]  & 0.0410 [0.0063] & 0.0440 [0.0065] & 0.0440 [0.0065] \\
\textbf{MCI prediction (4 classes)} & 0.0040 [0.0020] & 0.0010 [0.0010]  & 0.0010 [0.0010]  &0.0480 [0.0068] & 0.0330 [0.0056] & 0.0330 [0.0056] \\
\textbf{MCI prediction (2 classes)} & 0.0010 [0.0010] & 0.0010 [0.0010]  & 0.0010 [0.0010]  & - & - & - \\
\textbf{DIAN} & 0.0819 [0.0087] & 0.0020 [0.0014] & 0.0020 [0.0014] & 0.0300 [0.0054] & 0.0440 [0.0065] & 0.0440 [0.0065] 
\end{tabular}%
}
\medskip
The two cases associated with MCI prediction dataset results are due to the study of four-condition and two-condition distributions, respectively. The symbol ``-'' is used to express equality with the row above it. The level of significance was $0.05$ in all experiments.
\end{table*}

In general terms, the architectures generated through deep learning allow a high statistical power, since in most of the experiments the p-value was $0.0010$. Nevertheless, the null hypothesis of independence could not be rejected using CV in DIAN dataset (p-value=$0.0819$); $81$ out of the $1000$ actual errors associated with the null distribution were lower than the actual error of the original database. Thus, a tendency that statistical power is slightly lower using CV than RUB was observed.  This particularly occurred in cases where the accuracy rates related to the original datasets were not excessively high, as in the DIAN and the MCI prediction $4$-condition datasets. The main reason for this is that the test sets were smaller using $10$-fold CV than RUB, where the whole set was used as training and test set. Therefore, if the $1000$ values from $100$ complete $10$-fold iterations were considered, it is logical that a larger standard deviation appears than when using RUB. Furthermore, the computational cost was similar between $100$ iterations using CV and $1000$ using RUB, so in scenarios of low sample size and high computational cost (for example, as the configuration chosen for the ADNI), it is more convenient to use RUB with a low input dimensionality in the classifier.

Good control of false positives was also observed in all experiments, as the FWE rate was never greater than $0.05$. Nevertheless, if the standard deviation associated with the p-value was considered, where $68.27\%$ of the values associated with the p-value distribution would be located, its maximum could be greater than the $0.05$ significance level. This occurred using CV in the MCI prediction $4$-condition dataset and in the DIAN dataset using RUB. Previous studies \cite{Varoquaux2018} indicated the need for a high sample size in order to apply K-fold and obtain a low FWE, which is confirmed in this study since $100$, $123$ or $229$ are small sample sizes. For example, the highest FWE rate was $0.0480$ obtained in the case of having only $100$ samples (MCI prediction dataset).

A similarity between the accuracy distribution associated with CV test results and resubstitution in original datasets (statistical power studies) was observed in the three experiments (e.g. $0.6286$ and $0.6301$, respectively, in the case of the DIAN dataset). It should be noted that the results associated with the classification of the DIAN dataset were not excessively high due to the complexity of the data, as most of the subjects were young people with hardly any symptoms of Alzheimer's disease. Nonetheless, the values obtained were similar to those achieved in previous studies \cite{Castillo-Barnes2020}.  RUB accuracies were slightly lower than using resubstitution due to the upper bound correction, and therefore more conservative than using CV (e.g. $0.7656$ in the ADNI original dataset). In the case of permuted datasets, in both studies (statistical power and type I error) similarities were found between the mean values related to the test CV subset and RUB, with the standard deviation being smaller in the distributions obtained by RUB. For example, in the MCI prediction 2-condition dataset (statistical power study), the mean accuracy using RUB was $0.4913$ $[0.0263]$ and $0.5067$ $[0.4913]$ using CV. 

The discrepancy between the accuracy associated with the training subset and the test subset, or in other words, the generalization error, using CV can give an idea of the overfitting associated with the system. In the ADNI experiment, there was these values were similar in the original set ($0.8548$ for training and $0.8319$ for test). Furthermore, this proximity continued to occur when permuting the dataset, although with lower accuracies  closer to $50$\% ($0.4791$ and $0.4987$ for training and test subsets, respectively), which are logical results as data was permuted. In MCI prediction dataset, there was a greater separation between training and test-related results using CV (for example, $0.9141$ for training and $0.8458$ for test for the AD-HC original dataset).  The four-condition study had lower accuracy rates ($0.5823$ for training and $0.4396$ for test) as expected for a multiclass problem relative to a binary problem, with a $14\%$ difference between them. In the DIAN dataset, in the type I error study, an accuracy of almost $70$\% ($0.6830$) was reached in training using $10$-fold CV, which reinforces the difficulty of classifying this dataset, as differences were found even within the same subset of subjects without permutation. Furthermore, this training accuracy was far from the accuracy obtained in test subset ($0.5028$). Thus, a trend of larger separation between training and test results using CV was observed in one-condition design than in the two-condition design. The separation between training and test error was also observed in Figure \ref{fig:distribuciones_kfold}, from which it follows that deeper configurations, as that related to the ADNI dataset, can develop a greater capacity for generalization by increasing its complexity and learning capacity of the data. Thus, although it is a positive factor when the database is large and well labelled, one should be careful with the choice of data being trained as the results may be satisfactory when they should not be, detecting differences when they should be absent.

The influence of feature extraction on the results was tested by comparing the results obtained by applying the scheme illustrated in Figure \ref{fig:flow_clasificacion} and Figure \ref{fig:extraccion}. Although it did not influence the control of type I error or statistical power, a lower variability around $50\%$ was observed if feature extraction was performed before the permutation test, and not within the permutation test. There were two main reasons for this: i) feature extraction does not depend on the permuted database at any time, and ii) the variability associated with AE training is eliminated, since its optimisation is dependent on the samples' location in the set, among other factors. Nevertheless, the variability generated in the extraction of features is precisely one of the points that we wanted to highlight in this work. The scenario described in Section \ref{sec:reducida}, where a reduced dimensionality was used directly, does not represent  the current scenario in neuroscience, which is orientated more towards classifications using high complexity classifiers in complex spaces than on the use of linear SVM and low dimensional classifiers.

Lastly, a disadvantage of applying resubstitution with upper bound correction to analyse DL models is that the final classifier should be linear, as the computation of the upper bound becomes more complex in other circumstances. The large number of connections in the net and aspects to be considered such as activation functions or dropout make the application of the VC dimension in deep learning architectures too complex to generate upper bounds \cite{zhang2016}.

\section{Conclusion}
In this work, we compared the performance of cross-validation and upper-bounded resubstitution using permutation tests to obtain the residual scores, or errors, associated with deep learning architectures. Using the proposed methodology, the trade-off could be explored between statistical power and type I error by means of three different databases related to Alzheimer's disease. Furthermore,, overfitting can be detected by observing the discrepancy between the training and test errors using CV, as well as analysing the bound associated with this data.

As a conclusion, in general, the implemented models did not have sufficient effective capacity for fitting corrupted data with similar accuracy rates than with the original data. Nevertheless, this capacity will increase as the deep learning architecture becomes more sophisticated. Regarding capacity of generalization, the difference between training and test errors decreased as the complexity of the network increased. Moreover, this capacity of generalization was higher in experiments based on two-condition designs than in one-condition designs. The generalization error could be mitigated by using upper-bounded resubstitution as a validation method instead of cross-validation, as it was observed that the error related to RUB and the error associated with the test set in CV were similar using corrupted data. Furthermore, this similarity was also observed between resubstitution error and CV test set error using original data. This means that the CV test error is still related to RUB error, but the latter is less influenced by overfitting when a correction is applied. Both the CV and RUB methods  were associated with a false positive rate very close to the significance level for any input dimension. Additionally, both approaches offer acceptable statistical power, although it was slightly lower using CV. Thus, considering that the computational cost per iteration is lower using RUB than CV, we recommend its use for statistical studies.

\section*{Acknowledgments.} 
This work was partly supported by the Ministerio de Ciencia e Innovación (España)/ FEDER under the RTI2018-098913-B100 project, by the Consejería de Economía, Innovación, Ciencia y Empleo (Junta de Andalucía) and FEDER under CV20-45250 and A-TIC-080-UGR18 projects, and by Ministerio de Universidades (España) under the FPU predoctoral grant (FPU 18/04902) given to C. Jimenez-Mesa.

Data collection and sharing for this project was funded by the Alzheimer's Disease Neuroimaging Initiative (ADNI) (National Institutes of Health Grant U01 AG024904) and DOD ADNI (Department of Defense award number W81XWH-12-2-0012). ADNI is funded by the National Institute on Aging, the National Institute of Biomedical Imaging and Bioengineering, and through generous contributions from the following: AbbVie, Alzheimer’s Association; Alzheimer’s Drug Discovery Foundation; Araclon Biotech; BioClinica, Inc.; Biogen; Bristol-Myers Squibb Company; CereSpir, Inc.; Cogstate; Eisai Inc.; Elan Pharmaceuticals, Inc.; Eli Lilly and Company; EuroImmun; F. Hoffmann-La Roche Ltd and its affiliated company Genentech, Inc.; Fujirebio; GE Healthcare; IXICO Ltd.; Janssen Alzheimer Immunotherapy Research \& Development, LLC.; Johnson \& Johnson Pharmaceutical Research \& Development LLC.; Lumosity; Lundbeck; Merck \& Co., Inc.; Meso Scale Diagnostics, LLC.; NeuroRx Research; Neurotrack Technologies; Novartis Pharmaceuticals Corporation; Pfizer Inc.; Piramal Imaging; Servier; Takeda Pharmaceutical Company; and Transition Therapeutics. The Canadian Institutes of Health Research is providing funds to support ADNI clinical sites in Canada. Private sector contributions are facilitated by the Foundation for the National Institutes of Health (www.fnih.org). The grantee organization is the Northern California Institute for Research and Education, and the study is coordinated by the Alzheimer’s Therapeutic Research Institute at the University of Southern California. ADNI data are disseminated by the Laboratory for Neuro Imaging at the University of Southern California.

Data collection and sharing for this project was supported by The Dominantly Inherited Alzheimer’s Network (DIAN, U19AG032438) funded by the National Institute on Aging (NIA), the German Center for Neurodegenerative Diseases (DZNE), Raul Carrea Institute for Neurological Research (FLENI), Partial support by the Research and Development Grants for Dementia from Japan Agency for Medical Research and Development, AMED, and the Korea Health Technology R\&D Project through the Korea Health Industry Development Institute (KHIDI).This manuscript has been reviewed by DIAN Study investigators for scientific content and consistency of data interpretation with previous DIAN Study publications.  We acknowledge the altruism of the participants and their families and contributions of the DIAN research and support staff at each of the participating sites for their contributions to this study.

\section*{Appendix}
\subsection{Detailed database information}
\subsubsection{MCI prediction dataset} \label{appe:kaggle}
In the challenge, the dataset was divided in two different sets, one for training and other for testing with a total of $240$ ADNI real subjects in the training dataset ($60$ HC, $60$ MCI, $60$ cMCI and $60$ AD) and $500$ subjects in the testing one, where only 160 were real subjects. The remaining $340$ subjects were artificially generated from the real data. Demographic information of the real subjects is indicated in Tables \ref{tab:kaggle_train} and \ref{tab:kaggle_test}. 

Regarding challenge results, the winner team, which used a random forest ensemble, got a $61 \%$ accuracy in the multiclass classification problem \cite{dimitriadis_random_2018}. This result was outperformed after the challenge with $67\%$ by a one-versus-one based on error correcting output codes (ECOC) and SVM \cite{Jimenez-Mesa2020}. For more information on the challenge, the reader is advised to see the special issue \cite{sarica_editorial_2018} related to the results. 

\begin{table}[!h]

\caption{Training Dataset Information}\label{tab:kaggle_train}
\centering
\begin{tabular}{cccc}
N = 240 & \textbf{Gender (M/F)} & \textbf{Age} & \textbf{MMSE} \\
HC  &30/30  & 72.34 [5.67]  & 29.15 [1.11]\\
MCI  &28/32  & 72.19 [7.42] &  28.32 [1.56]\\
cMCI  &35/25  &72.96 [7.20]  & 27.18 [1.87]\\
AD & 29/31  &74.75 [7.31] &  23.43 [2.11]
\end{tabular}

\medskip
Sociodemographic data and MMSE for each group. $X [Y]$ denotes the mean $X$ and standard deviation $Y$ for each group
\end{table}

\begin{table}[!h]

\caption{Information of Real Data in Testing Dataset }\label{tab:kaggle_test}
\centering
\begin{tabular}{cccc}
N = 160 & \textbf{Gender (M/F)}  &\textbf{Age}  &\textbf{MMSE}\\
HC  &18/22 & 74.88 [5.48]  &29.00 [1.10]\\
MCI  &23/17 & 72.40 [8.04] & 27.65 [1.87]\\
cMCI & 25/15  &71.75 [6.23] & 27.58 [1.80]\\
AD  &23/17  &73.11  [8.05]  &22.68 [1.98]
\end{tabular}

\medskip
Sociodemographic data and MMSE for each group. $X [Y]$ denotes the mean $X$ and standard deviation $Y$ for each group
\end{table}

The original split between training set and test set associated to the MCI prediction dataset is preserved. The aim is to be able to compare the results obtained with those obtained in the challenge, and thus, be able to check whether the configuration chosen for the models works correctly.

\subsubsection{The DIAN dataset}\label{appe:dian}
The criteria for excluding subjects from the database for this study were as follows. Firstly, only data from the initial visit of the subjects is considered, which reduces the number of samples from $1219$ to $534$ samples. Then, $29$ patients were excluded as in \cite{Castillo-Barnes2020},  due to being diagnosed with at least one of the following diseases: cerebral stroke (3 subjects), transient ischemic attack (2 subject), dementia by alcoholism (4 subjects), Parkinson’s disease (1 subject), traumatic brain injury with chronic deficit/dysfunction (3 subjects), dementia with Lewy bodies (1 subject), vascular dementia (1 subject) and dementia by unknown causes (5 subjects). Besides, in order not to increase the heterogeneity in symptomatic subjects, Late Onset AD (LOAD) cases in the DIAN study have been also discarded (15 subjects). Finally, the criterion regarding missing values was to select those features that were fulfilled for at least $80\%$ of the subjects and to exclude the remaining features as well as incomplete subjects for the chosen features. Once this was done, the final set consists of $333$ samples with $722$ features each of them. The partition of subjects between NC and MC sets was $123$ subjects in the first group and $210$ in the MC group. Due to the large difference in the number of subjects in both groups, it was decided to balance them with $123$ in each group, reducing the set of MC subjects randomly.

\subsection{Upper-bounding}\label{appe:bound}
The upper bound can be seen as the difference between empirical and actual errors,  ${ \mu \geq \vert E_{act}\left( L_S(x) \right)  - E_{emp}\left( L_S(x) \right)   \vert }$. The upper bound commonly used in machine learning \cite{vapnik2013} is defined as: 
\begin{equation}\label{eq:vapnik}
\mu_{VC} \leq \sqrt{ \dfrac{h\left(   \ln{\left( \dfrac{2n}{h} \right) } + 1 \right) - \ln{\left( \dfrac{\eta}{4} \right) }  }   {n}  }
\end{equation}
where $\eta$ is the significance level, $n$ is the size of the training set, $d$ is the features dimension and $h$ is the VC dimension \cite{vapnik1994}. This dimension is equal to $d +1$ for linear functions, such as SVM, while for DL algorithms a lively debate exists about their value \cite{ zhang2016, bartlett2003, neyshabur2015} which is in any case proportional to the number of tunable parameters, $E$. In practice, since weights have a short representation as floating point numbers with $O(1)$ bits, the VC dimension is equal to $ O(\vert E\vert)$ \cite{shalev2014}. 

Another approach to estimate the upper bound for linear classifiers is tthe one described in the manuscript, equation (\ref{eq:gorriz}). It has been proved that $\mu_{emp}$ is tighter than the Vapnik' bound with regard to linear algorithms  \cite{Gorriz2019}. In both approaches, the high dependence of the bound value on the features dimension can be observed and why it is necessary to apply PLS to reduce the dimensionality with these estimates.

\subsection{Permutation test}\label{appe:perm_test}
The general procedure to perform a permutation test in a classification scenario is as follows:
\begin{enumerate}
\item Fix the test statistic $\mathcal{T}$ associated to the given the dataset $\mathbf{S} = \lbrace \left(  x_i, y_i  \right)  \rbrace_{i=1}^n  $. 
\item Generate a random permutation, $\pi$, of class labels.
\item Use the permuted dataset ${ \left\lbrace \mathbf{x},\mathbf{y}_{\pi}\right\rbrace  }$ for training and evaluating the classification algorithm.
\item Obtain the statistic $\mathcal{T_{\pi}}$.
\item Repeat (2.)-(4.) for $M$ iterations.
\item Given $\mathcal{T}$ and $\mathcal{T_{\pi}}$ distribution, calculate the p-value. 
\item Reject or not reject the null hypothesis. The rejection condition is to obtain a p-value lower than the level of significance, $\alpha$.
\end{enumerate}

\subsection{Deep learning theory}\label{appe:dl_teoria}
\subsection{Layers typology}
Fully-connected layers are those layers with $i$ perceptrons each one, where the connections go in forward direction and there are no connections within a layer. The former is the cause of one of the disadvantages associated with these layers, since as all perceptrons are connected with all those in the next layer, the number of parameters increases exponentially, which is inefficient. The equation associated with each perception is:
\begin{equation}\label{eq:neurona}
y_i^n = f\left( w_i^n \cdot \mathbf{y}^{n-1} + b_i^n  \right) 
\end{equation}
where $f(\cdot )$ is the activation function applied to the $i$-th perceptron of the layer $n$, $w_i^n$ is the the weight vector that multiplies the activations of the previous layer ($\mathbf{y}^{n-1}$) and $b_i^n$ is the associated bias.

Convolutional layers  have a higher complexity, allowing the application of other different algorithms (for example, pooling or transpose convolutions) to images, which results in outstanding results especially in image analysis and processing. Similar to equation (\ref{eq:neurona}), an equation for convolutional neurons can be posed:
\begin{equation}
y_i^n = f\left( w^n \cdot y_i^{n-1} + b_i^n  \right) 
\end{equation}
where only two differences exist, the $w^n$ term indicates that the weight matrix is shared by all the neurons in layer $n$, which allow the convolution, and $y_i^{n-1}$ means that for the neuron $i$ of layer $n$ only a part of the outputs of the previous layer is used, known as kernel size.

\subsubsection{Autoencoder}
An autoencoder is a neural network composed of two elements: a encoder $e(x)$ and a decoder $d(x)$. The former reduces the dimensionality of the data, while the latter recovers the initial dimensionality from the low-dimensionality data (Z-layer). This means that from a sample set of dimension $M$ such that $\mathbf{x_i} \in \mathbb{R}^M$, $e(\mathbf{x})$  allows to obtain a representation of these samples of dimension $Z$ such that $\mathbf{z_i} \in \mathbb{R}^Z$. The existence of this Z-layer make AE a relevant configuration as feature extractor, as it allows to associate input features with reduced dimensional space features without losing information \cite{Martinez-Murcia2020,baldi2012}.

Due to the nature of the autoencoder, the goal is to minimize the reconstruction error, which is the difference between the encoder input $\mathbf{x}$ and the decoder output $\mathbf{\hat{x}}=d\left( e\left( \mathbf{x}\right) \right) $. For the computation of this error, Mean Squared Error (MSE) algorithm is used:
\begin{equation}
MSE = \dfrac{1}{N} \sum_i \left( \mathbf{x_i} - d\left( e\left( \mathbf{x_i}\right) \right)\right) ^2
\end{equation}

\subsubsection{Activation functions}
A necessary element in a neural network is the activation function. Briefly, It defines the output of a node given the input, allowing non-linear transformations of the latter. The expressions of sigmoid and rectified linear unit (ReLU) activation functions are as follows:
\begin{equation}
\begin{split}
& f_{sigmoid}\left( x \right) = \dfrac{1}{1+\exp^{-x}}  \\
& f_{ReLU}\left( x \right) = max\left( 0, x \right) 
\end{split}
\end{equation}

\bibliographystyle{unsrt}

\begin{thebibliography}{10}

\bibitem{teipel_relative_2015}
Stefan~J. Teipel, Jens Kurth, Bernd Krause, and Michel~J. Grothe.
\newblock The relative importance of imaging markers for the prediction of
  {Alzheimer}'s disease dementia in mild cognitive impairment — {Beyond}
  classical regression.
\newblock {\em NeuroImage: Clinical}, 8:583--593, January 2015.

\bibitem{Martinez-Murcia2016}
F.~J. Martinez-Murcia, J.~M. Górriz, J.~Ramírez, A.~Ortiz, and for~the
  Alzheimer's Disease Neuroimaging~Initiative.
\newblock A spherical brain mapping of mr images for the detection of
  alzheimer's disease.
\newblock {\em Current Alzheimer Research}, 13(5):575--588, 2016.

\bibitem{khedher_independent_2016}
Laila Khedher, Ignacio~A. Illán, Juan~M. Górriz, Javier Ramírez, Abdelbasset
  Brahim, and Anke Meyer-Baese.
\newblock Independent {Component} {Analysis}-{Support} {Vector}
  {Machine}-{Based} {Computer}-{Aided} {Diagnosis} {System} for {Alzheimer}’s
  with {Visual} {Support}.
\newblock {\em International Journal of Neural Systems}, 27(03):1650050, July
  2016.

\bibitem{illan_spatial_2014}
Ignacio~A. Illan, Juan~M. Górriz, Javier Ramírez, and Anke Meyer-Base.
\newblock Spatial component analysis of {MRI} data for {Alzheimer}'s disease
  diagnosis: a {Bayesian} network approach.
\newblock {\em Frontiers in Computational Neuroscience}, 8:156, 2014.

\bibitem{segovia_comparative_2012}
F.~Segovia, J.M. Górriz, J.~Ramírez, D.~Salas-Gonzalez, I.~Álvarez,
  M.~López, and R.~Chaves.
\newblock A comparative study of feature extraction methods for the diagnosis
  of {Alzheimer}'s disease using the {ADNI} database.
\newblock {\em Neurocomputing}, 75(1):64--71, 2012.

\bibitem{davatzikos_machine_2019}
Christos Davatzikos.
\newblock Machine learning in neuroimaging: {Progress} and challenges.
\newblock {\em NeuroImage}, 197:652--656, August 2019.

\bibitem{Jo2019}
Taeho Jo, Kwangsik Nho, and Andrew~J. Saykin.
\newblock Deep learning in alzheimer's disease: Diagnostic classification and
  prognostic prediction using neuroimaging data.
\newblock {\em Frontiers in Aging Neuroscience}, 11, aug 2019.

\bibitem{martinez2018}
Francisco~J Martinez-Murcia, Juan~M G{\'o}rriz, Javier Ram{\'\i}rez, and Andres
  Ortiz.
\newblock Convolutional neural networks for neuroimaging in parkinson’s
  disease: is preprocessing needed?
\newblock {\em International journal of neural systems}, 28(10):1850035, 2018.

\bibitem{Ortiz2016}
Andr{\'{e}}s Ortiz, Jorge Munilla, Juan~M. G{\'{o}}rriz, and Javier
  Ram{\'{\i}}rez.
\newblock Ensembles of deep learning architectures for the early diagnosis of
  the alzheimer's disease.
\newblock {\em International Journal of Neural Systems}, 26(07):1650025, aug
  2016.

\bibitem{payan2015}
Adrien Payan and Giovanni Montana.
\newblock Predicting alzheimer's disease: a neuroimaging study with 3d
  convolutional neural networks.
\newblock {\em arXiv preprint arXiv:1502.02506}, 2015.

\bibitem{Ruck1990}
D.W. Ruck, S.K. Rogers, M.~Kabrisky, M.E. Oxley, and B.W. Suter.
\newblock The multilayer perceptron as an approximation to a bayes optimal
  discriminant function.
\newblock {\em {IEEE} Transactions on Neural Networks}, 1(4):296--298, 1990.

\bibitem{Wen2020}
Junhao Wen, Elina Thibeau-Sutre, Mauricio Diaz-Melo, Jorge
  Samper-Gonz{\'{a}}lez, Alexandre Routier, Simona Bottani, Didier Dormont,
  Stanley Durrleman, Ninon Burgos, and Olivier Colliot.
\newblock Convolutional neural networks for classification of alzheimer's
  disease: Overview and reproducible evaluation.
\newblock {\em Medical Image Analysis}, 63:101694, jul 2020.

\bibitem{Feng2020}
Wei Feng, Nicholas~Van Halm-Lutterodt, Hao Tang, Andrew Mecum, Mohamed~Kamal
  Mesregah, Yuan Ma, Haibin Li, Feng Zhang, Zhiyuan Wu, Erlin Yao, and Xiuhua
  Guo.
\newblock Automated {MRI}-based deep learning model for detection of
  alzheimer's disease process.
\newblock {\em International Journal of Neural Systems}, 30(06):2050032, may
  2020.

\bibitem{ronneberger2015u}
Olaf Ronneberger, Philipp Fischer, and Thomas Brox.
\newblock U-net: Convolutional networks for biomedical image segmentation.
\newblock In {\em International Conference on Medical image computing and
  computer-assisted intervention}, pages 234--241. Springer, 2015.

\bibitem{Martinez-Murcia2020}
Francisco~J. Martinez-Murcia, Andres Ortiz, Juan-Manuel Gorriz, Javier Ramirez,
  and Diego Castillo-Barnes.
\newblock Studying the manifold structure of alzheimer's disease: A deep
  learning approach using convolutional autoencoders.
\newblock {\em {IEEE} Journal of Biomedical and Health Informatics},
  24(1):17--26, jan 2020.

\bibitem{Basu2019}
Sumana Basu, Konrad Wagstyl, Azar Zandifar, Louis Collins, Adriana Romero, and
  Doina Precup.
\newblock Early prediction of alzheimer's disease progression using variational
  autoencoders.
\newblock In {\em Lecture Notes in Computer Science}, pages 205--213. Springer
  International Publishing, 2019.

\bibitem{Suk2013}
Heung-Il Suk, , Seong-Whan Lee, and Dinggang Shen.
\newblock Latent feature representation with stacked auto-encoder for
  {AD}/{MCI} diagnosis.
\newblock {\em Brain Structure and Function}, 220(2):841--859, dec 2013.

\bibitem{Weiner2013}
Michael~W. Weiner, Dallas~P. Veitch, Paul~S. Aisen, Laurel~A. Beckett, Nigel~J.
  Cairns, Robert~C. Green, Danielle Harvey, Clifford~R. Jack, William Jagust,
  Enchi Liu, John~C. Morris, Ronald~C. Petersen, Andrew~J. Saykin, Mark~E.
  Schmidt, Leslie Shaw, Li~Shen, Judith~A. Siuciak, Holly Soares, Arthur~W.
  Toga, and John Q.~Trojanowski and.
\newblock The alzheimer's disease neuroimaging initiative: A review of papers
  published since its inception.
\newblock {\em Alzheimer's {\&} Dementia}, 9(5):e111--e194, jul 2013.

\bibitem{Morris2012}
John~C Morris, Paul~S Aisen, Randall~J Bateman, Tammie~LS Benzinger, Nigel~J
  Cairns, Anne~M Fagan, Bernardino Ghetti, Alison~M Goate, David~M Holtzman,
  William~E Klunk, Eric McDade, Daniel~S Marcus, Ralph~N Martins, Colin~L
  Masters, Richard Mayeux, Angela Oliver, Kimberly Quaid, John~M Ringman,
  Martin~N Rossor, Stephen Salloway, Peter~R Schofield, Natalie~J Selsor,
  Reisa~A Sperling, Michael~W Weiner, Chengjie Xiong, Krista~L Moulder, and
  Virginia~D Buckles.
\newblock Developing an international network for alzheimer's research: the
  dominantly inherited alzheimer network.
\newblock {\em Clinical Investigation}, 2(10):975--984, oct 2012.

\bibitem{Neu2016}
Scott~C. Neu, Karen~L. Crawford, and Arthur~W. Toga.
\newblock Sharing data in the global alzheimer's association interactive
  network.
\newblock {\em {NeuroImage}}, 124:1168--1174, jan 2016.

\bibitem{Castiglioni2018}
Isabella Castiglioni, Christian Salvatore, Javier Ram{\'{\i}}rez, and
  Juan~Manuel G{\'{o}}rriz.
\newblock Machine-learning neuroimaging challenge for automated diagnosis of
  mild cognitive impairment: Lessons learnt.
\newblock {\em Journal of Neuroscience Methods}, 302:10--13, may 2018.

\bibitem{kawaguchi2017}
Kenji Kawaguchi, Leslie~Pack Kaelbling, and Yoshua Bengio.
\newblock Generalization in deep learning.
\newblock {\em arXiv preprint arXiv:1710.05468}, 2017.

\bibitem{neyshabur2017}
Behnam Neyshabur, Srinadh Bhojanapalli, David McAllester, and Nati Srebro.
\newblock Exploring generalization in deep learning.
\newblock {\em Advances in neural information processing systems},
  30:5947--5956, 2017.

\bibitem{srivastava2014}
Nitish Srivastava, Geoffrey Hinton, Alex Krizhevsky, Ilya Sutskever, and Ruslan
  Salakhutdinov.
\newblock Dropout: a simple way to prevent neural networks from overfitting.
\newblock {\em The journal of machine learning research}, 15(1):1929--1958,
  2014.

\bibitem{ioffe2015}
Sergey Ioffe and Christian Szegedy.
\newblock Batch normalization: Accelerating deep network training by reducing
  internal covariate shift.
\newblock {\em arXiv preprint arXiv:1502.03167}, 2015.

\bibitem{Shorten2019}
Connor Shorten and Taghi~M. Khoshgoftaar.
\newblock A survey on image data augmentation for deep learning.
\newblock {\em Journal of Big Data}, 6(1), jul 2019.

\bibitem{Thanapol2020}
Panissara Thanapol, Kittichai Lavangnananda, Pascal Bouvry, Frederic Pinel, and
  Franck Leprevost.
\newblock Reducing overfitting and improving generalization in training
  convolutional neural network ({CNN}) under limited sample sizes in image
  recognition.
\newblock In {\em 2020 - 5th International Conference on Information Technology
  ({InCIT})}. {IEEE}, oct 2020.

\bibitem{Golland2005}
Polina Golland, Feng Liang, Sayan Mukherjee, and Dmitry Panchenko.
\newblock Permutation tests for classification.
\newblock In {\em Learning Theory}, pages 501--515. Springer Berlin Heidelberg,
  2005.

\bibitem{Ojala2009}
Markus Ojala and Gemma~C. Garriga.
\newblock Permutation tests for studying classifier performance.
\newblock In {\em 2009 Ninth {IEEE} International Conference on Data Mining}.
  {IEEE}, dec 2009.

\bibitem{Pereira2009}
Francisco Pereira, Tom Mitchell, and Matthew Botvinick.
\newblock Machine learning classifiers and {fMRI}: A tutorial overview.
\newblock {\em {NeuroImage}}, 45(1):S199--S209, mar 2009.

\bibitem{Golland2003}
Polina Golland and Bruce Fischl.
\newblock Permutation tests for classification: Towards statistical
  significance in image-based studies.
\newblock In {\em Lecture Notes in Computer Science}, pages 330--341. Springer
  Berlin Heidelberg, 2003.

\bibitem{Rosenblatt2019}
Jonathan~D Rosenblatt, Yuval Benjamini, Roee Gilron, Roy Mukamel, and Jelle~J
  Goeman.
\newblock Better-than-chance classification for signal detection.
\newblock {\em Biostatistics}, oct 2019.

\bibitem{Etzel2009}
Joset~A. Etzel, Valeria Gazzola, and Christian Keysers.
\newblock An introduction to anatomical {ROI}-based {fMRI} classification
  analysis.
\newblock {\em Brain Research}, 1282:114--125, jul 2009.

\bibitem{Stelzer2013}
Johannes Stelzer, Yi~Chen, and Robert Turner.
\newblock Statistical inference and multiple testing correction in
  classification-based multi-voxel pattern analysis ({MVPA}): Random
  permutations and cluster size control.
\newblock {\em {NeuroImage}}, 65:69--82, jan 2013.

\bibitem{Gorriz2021}
J.M. Gorriz, C.~Jimenez-Mesa, R.~Romero-Garcia, F.~Segovia, J.~Ramirez,
  D.~Castillo-Barnes, F.J. Martinez-Murcia, A.~Ortiz, D.~Salas-Gonzalez, I.A.
  Illan, C.G. Puntonet, D.~Lopez-Garcia, M.~Gomez-Rio, and J.~Suckling.
\newblock Statistical agnostic mapping: A framework in neuroimaging based on
  concentration inequalities.
\newblock {\em Information Fusion}, 66:198--212, feb 2021.

\bibitem{Good2013}
Phillip Good.
\newblock {\em Permutation tests: a practical guide to resampling methods for
  testing hypotheses}.
\newblock Springer Science \& Business Media, 2013.

\bibitem{Eklund2016}
Anders Eklund, Thomas~E. Nichols, and Hans Knutsson.
\newblock Cluster failure: Why {fMRI} inferences for spatial extent have
  inflated false-positive rates.
\newblock {\em Proceedings of the National Academy of Sciences},
  113(28):7900--7905, jun 2016.

\bibitem{Friston1994}
K.~J. Friston, A.~P. Holmes, K.~J. Worsley, J.-P. Poline, C.~D. Frith, and
  R.~S.~J. Frackowiak.
\newblock Statistical parametric maps in functional imaging: A general linear
  approach.
\newblock {\em Human Brain Mapping}, 2(4):189--210, 1994.

\bibitem{Ashburner2005}
John Ashburner and Karl~J. Friston.
\newblock Unified segmentation.
\newblock {\em {NeuroImage}}, 26(3):839--851, jul 2005.

\bibitem{fischl_measuring_2000}
B.~Fischl and A.~M. Dale.
\newblock Measuring the thickness of the human cerebral cortex from magnetic
  resonance images.
\newblock {\em Proceedings of the National Academy of Sciences},
  97(20):11050--11055, sep 2000.

\bibitem{fischl_freesurfer_2012}
Bruce Fischl.
\newblock {FreeSurfer}.
\newblock {\em NeuroImage}, 62(2):774--781, August 2012.

\bibitem{Levy1990}
E.~Levy, M.~Carman, I.~Fernandez-Madrid, M.~Power, I.~Lieberburg, S.~van
  Duinen, G.~Bots, W.~Luyendijk, and B.~Frangione.
\newblock Mutation of the alzheimer's disease amyloid gene in hereditary
  cerebral hemorrhage, dutch type.
\newblock {\em Science}, 248(4959):1124--1126, jun 1990.

\bibitem{Ryan2016}
Natalie~S Ryan, Jennifer~M Nicholas, Philip S~J Weston, Yuying Liang, Tammaryn
  Lashley, Rita Guerreiro, Gary Adamson, Janna Kenny, Jon Beck, Lucia
  Chavez-Gutierrez, Bart de~Strooper, Tamas Revesz, Janice Holton, Simon Mead,
  Martin~N Rossor, and Nick~C Fox.
\newblock Clinical phenotype and genetic associations in autosomal dominant
  familial alzheimer's disease: a case series.
\newblock {\em The Lancet Neurology}, 15(13):1326--1335, dec 2016.

\bibitem{Levy-Lahad1995}
E.~Levy-Lahad, W.~Wasco, P.~Poorkaj, D.~Romano, J.~Oshima, W.~Pettingell,
  C.~Yu, P.~Jondro, S.~Schmidt, K.~Wang, and e.~al.
\newblock Candidate gene for the chromosome 1 familial alzheimer's disease
  locus.
\newblock {\em Science}, 269(5226):973--977, aug 1995.

\bibitem{Morris1997}
John~C. Morris.
\newblock Clinical dementia rating: A reliable and valid diagnostic and staging
  measure for dementia of the alzheimer type.
\newblock {\em International Psychogeriatrics}, 9(S1):173--176, dec 1997.

\bibitem{Berg1988}
L.~Berg.
\newblock Clinical dementia rating (cdr).
\newblock {\em Psychopharmacol Bull}, (24):637--639, 1988.

\bibitem{Fagan2014}
A.~M. Fagan, C.~Xiong, M.~S. Jasielec, R.~J. Bateman, A.~M. Goate, T.~L.~S.
  Benzinger, B.~Ghetti, R.~N. Martins, C.~L. Masters, R.~Mayeux, J.~M. Ringman,
  M.~N. Rossor, S.~Salloway, P.~R. Schofield, R.~A. Sperling, D.~Marcus, N.~J.
  Cairns, V.~D. Buckles, J.~H. Ladenson, J.~C. Morris, and D.~M. Holtzman.
\newblock Longitudinal change in {CSF} biomarkers in autosomal-dominant
  alzheimer's disease.
\newblock {\em Science Translational Medicine}, 6(226):226ra30--226ra30, mar
  2014.

\bibitem{vapnik1982}
VN~Vapnik.
\newblock Estimation of dependencies based on empirical data springer.
\newblock {\em Information and Control}, 1982.

\bibitem{Gorriz2019}
Juan~M. Górriz, Javier Ramirez, and John Suckling.
\newblock On the computation of distribution-free performance bounds:
  Application to small sample sizes in neuroimaging.
\newblock {\em Pattern Recognition}, 93:1--13, sep 2019.

\bibitem{Reiss2015}
Philip~T. Reiss.
\newblock Cross-validation and hypothesis testing in neuroimaging: An irenic
  comment on the exchange between friston and lindquist et al.
\newblock {\em {NeuroImage}}, 116:248--254, aug 2015.

\bibitem{Efron1993}
Bradley Efron and Robert~J. Tibshirani.
\newblock {\em An Introduction to the Bootstrap}.
\newblock Springer {US}, 1993.

\bibitem{Lopez2009}
M.M. L{\'{o}}pez, J.~Ram{\'{\i}}rez, J.M. G{\'{o}}rriz, I.~{\'{A}}lvarez,
  D.~Salas-Gonzalez, F.~Segovia, and R.~Chaves.
\newblock {SVM}-based {CAD} system for early detection of the alzheimer's
  disease using kernel {PCA} and {LDA}.
\newblock {\em Neuroscience Letters}, 464(3):233--238, oct 2009.

\bibitem{Jimenez-Mesa2020}
Carmen Jimenez-Mesa, Ignacio~Alvarez Illan, Alberto Martin-Martin, Diego
  Castillo-Barnes, Francisco~Jesus Martinez-Murcia, Javier Ramirez, and Juan~M.
  Gorriz.
\newblock Optimized one vs one approach in multiclass classification for early
  alzheimer's disease and mild cognitive impairment diagnosis.
\newblock {\em {IEEE} Access}, 8:96981--96993, 2020.

\bibitem{Gorriz2017}
J.~M. Gorriz, J.~Ramirez, J.~Suckling, Ignacio~Alvarez Illan, Andres Ortiz,
  F.~J. Martinez-Murcia, Fermin Segovia, D.~Salas-Gonzalez, and Shuihua Wang.
\newblock Case-based statistical learning: A non-parametric implementation with
  a conditional-error rate {SVM}.
\newblock {\em {IEEE} Access}, 5:11468--11478, 2017.

\bibitem{Ghazal2018}
Mohammed Ghazal.
\newblock Alzheimer's disease diagnostics by a 3d deeply supervised adaptable
  convolutional network.
\newblock {\em Frontiers in Bioscience}, 23(2):584--596, 2018.

\bibitem{rosipal2005}
Roman Rosipal and Nicole Kr{\"a}mer.
\newblock Overview and recent advances in partial least squares.
\newblock In {\em International Statistical and Optimization Perspectives
  Workshop" Subspace, Latent Structure and Feature Selection"}, pages 34--51.
  Springer, 2005.

\bibitem{kingma2019}
Diederik~P Kingma and J~Adam Ba.
\newblock A method for stochastic optimization. arxiv 2014.
\newblock {\em arXiv preprint arXiv:1412.6980}, 434, 2019.

\bibitem{Tzourio-Mazoyer2002}
N.~Tzourio-Mazoyer, B.~Landeau, D.~Papathanassiou, F.~Crivello, O.~Etard,
  N.~Delcroix, B.~Mazoyer, and M.~Joliot.
\newblock Automated anatomical labeling of activations in {SPM} using a
  macroscopic anatomical parcellation of the {MNI} {MRI} single-subject brain.
\newblock {\em {NeuroImage}}, 15(1):273--289, jan 2002.

\bibitem{Castillo-Barnes2020}
Diego Castillo-Barnes, Li~Su, Javier Ram{\'{\i}}rez, Diego Salas-Gonzalez,
  Francisco~J. Martinez-Murcia, Ignacio~A. Illan, Fermin Segovia, Andres Ortiz,
  Carlos Cruchaga, Martin~R. Farlow, Chengjie Xiong, Neil~R. Graff-Radford,
  Peter~R. Schofield, Colin~L. Masters, Stephen Salloway, Mathias Jucker,
  Hiroshi Mori, Johannes Levin, Juan~M. Gorriz, and Dominantly Inherited
  Alzheimer~Network (DIAN).
\newblock Autosomal dominantly inherited alzheimer disease: Analysis of genetic
  subgroups by machine learning.
\newblock {\em Information Fusion}, 58:153--167, jun 2020.

\bibitem{Franzmeier2020}
Nicolai Franzmeier, Nikolaos Koutsouleris, Tammie Benzinger, Alison Goate,
  Celeste~M. Karch, Anne~M. Fagan, Eric McDade, Marco Duering, Martin Dichgans,
  Johannes Levin, Brian~A. Gordon, Yen~Ying Lim, Colin~L. Masters, Martin
  Rossor, Nick~C. Fox, Antoinette O'Connor, Jasmeer Chhatwal, Stephen Salloway,
  Adrian Danek, Jason Hassenstab, Peter~R. Schofield, John~C. Morris,
  Randall~J. Bateman, Michael Ewers, and and.
\newblock Predicting sporadic alzheimer's disease progression via inherited
  alzheimer's disease-informed machine-learning.
\newblock {\em Alzheimer's {\&} Dementia}, 16(3):501--511, feb 2020.

\bibitem{Luckett2021}
Patrick~H. Luckett, Austin McCullough, Brian~A. Gordon, Jeremy Strain, Shaney
  Flores, Aylin Dincer, John McCarthy, Todd Kuffner, Ari Stern, Karin~L.
  Meeker, Sarah~B. Berman, Jasmeer~P. Chhatwal, Carlos Cruchaga, Anne~M. Fagan,
  Martin~R. Farlow, Nick~C. Fox, Mathias Jucker, Johannes Levin, Colin~L.
  Masters, Hiroshi Mori, James~M. Noble, Stephen Salloway, Peter~R. Schofield,
  Adam~M. Brickman, William~S. Brooks, David~M. Cash, Michael~J. Fulham,
  Bernardino Ghetti, Clifford~R. Jack, Jonathan Vöglein, William Klunk, Robert
  Koeppe, Hwamee Oh, Yi~Su, Michael Weiner, Qing Wang, Laura Swisher, Dan
  Marcus, Deborah Koudelis, Nelly Joseph-Mathurin, Lisa Cash, Russ Hornbeck,
  Chengjie Xiong, Richard~J. Perrin, Celeste~M. Karch, Jason Hassenstab, Eric
  McDade, John~C. Morris, Tammie~L.S. Benzinger, Randall~J. Bateman, and Beau
  M.~Ances and.
\newblock Modeling autosomal dominant alzheimer's disease with machine
  learning.
\newblock {\em Alzheimer's {\&} Dementia}, jan 2021.

\bibitem{Abdi2010}
Herv{\'{e}} Abdi and Lynne~J. Williams.
\newblock Principal component analysis.
\newblock {\em Wiley Interdisciplinary Reviews: Computational Statistics},
  2(4):433--459, jun 2010.

\bibitem{Varoquaux2018}
Gaël Varoquaux.
\newblock Cross-validation failure: Small sample sizes lead to large error
  bars.
\newblock {\em {NeuroImage}}, 180:68--77, oct 2018.

\bibitem{zhang2016}
Chiyuan Zhang, Samy Bengio, Moritz Hardt, Benjamin Recht, and Oriol Vinyals.
\newblock Understanding deep learning requires rethinking generalization.
\newblock {\em arXiv preprint arXiv:1611.03530}, 2016.

\bibitem{dimitriadis_random_2018}
S.~I. Dimitriadis, Dimitris Liparas, and Magda~N. Tsolaki.
\newblock Random forest feature selection, fusion and ensemble strategy:
  {Combining} multiple morphological {MRI} measures to discriminate among
  healhy elderly, {MCI}, {cMCI} and alzheimer’s disease patients: {From} the
  alzheimer’s disease neuroimaging initiative ({ADNI}) database.
\newblock {\em Journal of Neuroscience Methods}, 302:14--23, May 2018.

\bibitem{sarica_editorial_2018}
Alessia Sarica, Antonio Cerasa, Aldo Quattrone, and Vince Calhoun.
\newblock Editorial on special issue: {Machine} learning on {MCI}.
\newblock {\em Journal of Neuroscience Methods}, 302:1--2, May 2018.

\bibitem{vapnik2013}
Vladimir Vapnik.
\newblock {\em The nature of statistical learning theory}.
\newblock Springer science \& business media, 2013.

\bibitem{vapnik1994}
Vladimir Vapnik, Esther Levin, and Yann~Le Cun.
\newblock Measuring the vc-dimension of a learning machine.
\newblock {\em Neural computation}, 6(5):851--876, 1994.

\bibitem{bartlett2003}
Peter~L. Bartlett and Wolfgang Maass.
\newblock Vapnik-chervonenkis dimension of neural nets.
\newblock {\em The handbook of brain theory and neural networks}, pages
  1188--1192, 2003.

\bibitem{neyshabur2015}
Behnam Neyshabur, Ryota Tomioka, and Nathan Srebro.
\newblock Norm-based capacity control in neural networks.
\newblock In {\em Conference on Learning Theory}, pages 1376--1401, 2015.

\bibitem{shalev2014}
Shai Shalev-Shwartz and Shai Ben-David.
\newblock {\em Understanding machine learning: From theory to algorithms}.
\newblock Cambridge university press, 2014.

\bibitem{baldi2012}
Pierre Baldi.
\newblock Autoencoders, unsupervised learning, and deep architectures.
\newblock In {\em Proceedings of ICML workshop on unsupervised and transfer
  learning}, pages 37--49, 2012.


\end{thebibliography}

\end{document}